\documentclass[journal=jacsat,manuscript=article]{achemso}
\mciteErrorOnUnknownfalse
\usepackage{graphicx}
\usepackage{subcaption}

\usepackage[version=3]{mhchem} 
\usepackage{siunitx}
\usepackage[dvipsnames]{xcolor}
\usepackage{amssymb}
\usepackage{pifont}
\usepackage{wrapfig}
\usepackage{booktabs}
\usepackage{multirow}
\usepackage{hyperref}
\usepackage{microtype}
\usepackage{geometry}
\usepackage{array}
\usepackage{makecell}
\usepackage{colortbl}
\usepackage{algorithm}
\usepackage{algpseudocode}

\usepackage{amsmath} 

\newcommand{\iu}{\mathrm{i}\mkern1mu}

\SectionsOn
\SectionNumbersOn
\author{Zijie Li}
\affiliation[MechE]{Department of Mechanical Engineering,
Carnegie Mellon University,
Pittsburgh PA, USA}
\alsoaffiliation{Contributed equally to this work}

\author{Saurabh Patil}
\affiliation[MechE]{Department of Mechanical Engineering,
Carnegie Mellon University,
Pittsburgh PA, USA}
\alsoaffiliation{Contributed equally to this work}

\author{Francis Ogoke}
\affiliation[MechE]{Department of Mechanical Engineering,
Carnegie Mellon University,
Pittsburgh PA, USA}
\alsoaffiliation{Contributed equally to this work}

\author{Dule Shu}
\affiliation[MechE]{Department of Mechanical Engineering,
Carnegie Mellon University,
Pittsburgh PA, USA}

\author{Wilson Zhen}
\affiliation[MechE]{Department of Mechanical Engineering,
Carnegie Mellon University,
Pittsburgh PA, USA}

\author{\\Michael Schneier}
\affiliation[NNL]{Naval Nuclear Laboratory, West Mifflin PA, USA}

\author{John R. Buchanan, Jr.}
\affiliation[NNL]{Naval Nuclear Laboratory, West Mifflin PA, USA}

\author{Amir Barati Farimani}
\affiliation[MechE]{Department of Mechanical Engineering,
Carnegie Mellon University,
Pittsburgh PA, USA}
\email{barati@cmu.edu}

\title
  {Latent Neural PDE Solver: a reduced-order modelling framework for partial differential equations}
\abbreviations{IR,NMR,UV}

\begin{document}









\begin{abstract}
  Neural networks have shown promising potential in accelerating the numerical simulation of systems governed by partial differential equations (PDEs). Different from many existing neural network surrogates operating on high-dimensional discretized fields, we propose to learn the dynamics of the system in the latent space with much coarser discretizations. In our proposed framework - Latent Neural PDE Solver (LNS), a non-linear autoencoder is first trained to project the full-order representation of the system onto the mesh-reduced space, then a temporal model is trained to predict the future state in this mesh-reduced space. This reduction process simplifies the training of the temporal model by greatly reducing the computational cost accompanying a fine discretization and enables more efficient backprop-through-time training. We study the capability of the proposed framework and several other popular neural PDE solvers on various types of systems including single-phase and multi-phase flows along with varying system parameters. We showcase that it has competitive accuracy and efficiency compared to the neural PDE solver that operates on full-order space.
\end{abstract}

\section{Introduction}
Many intricate physical processes, from the interaction of protein dynamics to the movement of a celestial body, can be described by time-dependent partial differential equations (PDEs). The simulation of these processes is often conducted by solving these equations numerically, which requires fine discretization to resolve the necessary spatiotemporal domain to reach convergence. Deep neural network surrogates \citep{raissi2019physics, li2020fourier, lu2019deeponet, sanchez2020learning, pfaff2021learning, brandstetter2022message}  recently emerged as a computationally less-expensive alternative, with the potential to improve the efficiency of simulation by relaxing the requirement for fine discretization and attaining a higher accuracy on coarser grids compared to classical numerical solver \citep{stachenfeld2022learned, li2020fourier, brandstetter2022message}. 

For time-dependent systems, many neural-network-based models address the problem by approximating the solution operator $\mathcal{G}$ that maps the state $u_t$ to $u_{t+\Delta t}$, where the input and output are sampled on discretization grid $\{D_i, D_h\}$ respectively. The input discretization grid can either remain unchanged between every layer inside the network \citep{li2020fourier, brandstetter2022message, cao2021galerkin}, or fit into a hierarchical structure \citep{gupta2023towards, li2020multipole, wang2020towards, rahman2023uno, thuerey2020deep, Pant2021dlrom} that resembles the V-Cycle in classical multi-grid methods. Hierarchical structures have also been a common model architectural choice in the field of image segmentation \citep{ronneberger2015unet} and generation \citep{ho2020denoising} given their capability for utilizing multi-scale information. 

In contrast to the aforementioned approaches, especially those that utilize a hierarchical grid structure, our work studies the effect of decoupling dynamics prediction from upsampling/downsampling processes. Specifically, the neural network for predicting the forward dynamics (which we defined as a propagator) only operates on the coarsest resolution, while a deep autoencoder is pre-trained to compress the data from the original discretization grid $D_i$ to the coarse grid $D_l$ (e.g. from a $64\times64$ grid to an $8\times8$ grid). As the propagator network operates on a lower dimensional space, the training cost is greatly reduced and can be potentially adapted to unrolled training with a longer backprop-through-time (BPTT) horizon, which is often observed to be helpful to long-term stability \citep{han2022predicting, geneva2022transformers, list2024temporal}. We parameterize the model with a convolutional neural network along with several other components that are popular in neural PDE solvers, including several variants of attention. We test the proposed framework on different time-dependent PDEs with different flow types and boundary conditions. We showcase that the model can achieve efficient data compression and accurate prediction of complicated forward dynamics with significantly reduction in the computational cost associated with backpropagation through time. The proposed framework highlights an alternative for choosing the space for neural PDE solver to operate on.

\section{Related works}
\paragraph{Neural PDE solver}
Neural PDE solvers can be categorized into the following groups based on their model design. The first group employs neural networks with mesh-specific architectures, such as convolutional layers for uniform meshes or graph layers for irregular meshes. These networks learn spatiotemporal correlations within PDE data without the knowledge of the underlying equations \citep{gupta2023towards, stachenfeld2022learned, brandstetter2022message, sanchez2020learning, pfaff2021learning, Ummenhofer2020Lagrangian, prantl2022guaranteed, li2022fgn, lotzsch2022learning, Pant2021dlrom, wang2020towards, li2019dpi, thuerey2020deep, janny2023eagle}. Such a data-driven approach is useful for systems with unknown or partially known physics, such as large-scale climate modeling \citep{Rasp2020weatherbench, nguyen2023climax, lam2022graphcast, pathak2022fourcastnet}. The second group, known as Physics-Informed Neural Networks (PINNs) \citep{raissi2019physics, zhu2023reliable, pang2019fpinns, lu2021deepxde, cai2021physics, karniadakis2021physics}, treats neural networks as a parameterization of the solution function. PINNs incorporate knowledge of governing equations into the loss function, including PDE residuals and consistency with boundary and initial conditions. Unlike the first group, PINNs can be trained solely on equation loss and do not necessarily require input-target data pairs. The third group, known as the neural operators \citep{kissas2022learning, kovachki2021neural, brandstetter2022clifford, li2020neural, brandstetter2022lipoint, li2020fourier, li2020multipole, lu2019deeponet, jin2022mionet, gupta2021multiwaveletbased, cao2021galerkin, li2023transformer, hao2023gnot, ovadia2023vito, Lu2022fair, bhattacharya2021pcanet}, is designed to learn the mapping between function spaces. For a certain family of PDEs, neural operators have the potential to generalize and adapt to multiple discretizations. DeepONet \citep{lu2019deeponet} presents a pragmatic implementation of the universal operator approximation theorem\citep{Universal-apprx-operator-IEEE-1995}. Meanwhile, the concurrent research \citep{li2020gno} in the form of the graph neural operator proposes a trainable kernel integral for approximating solution operators in parametric PDEs. Their follow-up work, Fourier Neural Operator (FNO) \citep{li2020fourier}, has demonstrated high accuracy and efficiency in solving specific types of problems. Different function bases such as Fourier\citep{li2020fourier, wen2022ufno, tran2023factfno, guibas2021adaptive} / wavelet bases\citep{gupta2021multiwaveletbased}, the learned feature map from attention layers\citep{cao2021galerkin, li2023transformer}, or Green's function approximation\citep{boulle2022learning, tang2022neural}, have been be used for operator learning. For more physically consistent predictions, neural operator training can be combined with PINN principles \citep{li2023physicsinformed, pi-deeponet, lorsung2024picl, lorsung2024physics}.

\paragraph{Two-stage model for image compression and synthesis}
The utilization of a two-stage model for image synthesis has gained significant attention in the field of computer vision in recent years. Vector Quantized Variational Autoencoders (VQ-VAEs)\citep{razavi2019generating} adopt a two-stage approach for generating images within a latent space. In the initial stage, the approach compresses images into this latent space, using model components such as an encoder, a codebook, and a decoder. Subsequently, in the second stage, a latent model is introduced to predict the latent characteristics of the compressed images, and the decoder from the first stage is used to transform the predicted latent representation back into image pixels. Vector Quantized Generative Adversarial Networks (VQ-GANs)\citep{esser2021taming} are developed to scale autoregressive transformers to large image generation by employing adversarial and perceptual objectives for first-stage training. Most recently, several works have developed latent diffusion models with promising results ranging from image\citep{rombach2022highresolution} to point clouds\citep{zeng2022lion}.

\paragraph{Latent space modelling for time-dependent PDE problems}
Many of the existing time-dependent neural PDE solvers have employed Encoder-Process-Decoder (EPD) scheme, used to map the input solution at time $t$ to the subsequent time step \citep{brandstetter2022message, Coarse-Turbulence-ICLR-2022, pfaff2021learning, sanchez2020learning, PDE-ROM-2021,hsieh-AR-ICLR-2019}. In these models, the input mesh and output mesh are usually the same, where the encoder and decoder are pointwise operator that does not affect the underlying discretization. As an alternative, researchers have explored propagating the system dynamics in the latent space with coarsened mesh, aiming to reduce computational complexity and minimize memory usage \citep{pod-dl-rom, deep-conservation-latent-AAAI-2021, Evolution-fluid-CG-2019, vlachas2022led, wang2024lno, kontolati2024latentdeeponet}.Evolving the system dynamics in latent space can involve utilization of recurrent neural networks like Long Short-Term Memory \citep{Evolution-fluid-CG-2019}, linear propagators grounded in the assumptions of the Koopman operator \citep{Dyn-modeling-koopman-NIPS-2018, Compositional-Koopman-ICLR-2020, Deep-Koopman-Nature-2018, Koopman-DMD-NIPS-2017, Koopman-Stability-SIAM-2020}, reduced basis space obtained through neural networks \citep{fresca2021comprehensive-rom} or a combination of proper orthogonal decomposition and neural networks \citep{pod-dl-rom}, more recently spatial attention mechanism \citep{hemmasian2023reduced, hemmasian2023multiscale} and temporal attention mechanism \citep{han2022predicting}, recurrent Multilayer Perceptrons \citep{li2023transformer}, state-space model \citep{patil2023hno} or continuous dynamics model \citep{yin2023continuouspde, rojas2021reduced} based on neural Ordinary Equations (ODEs) \citep{neuralode2018}. In this work, we propose to first train an autoencoder to embed and compress inputs into the latent space, and then employ a Markovian propagator to learn the dynamics of the time-dependent system within this latent space (Figure \ref{fig:schematic}). A detailed schematic for the model architecture used in this work is provided in the Appendix Figure 8.

\begin{figure}[H]
    \centering   
    \includegraphics[width=1.0\textwidth]{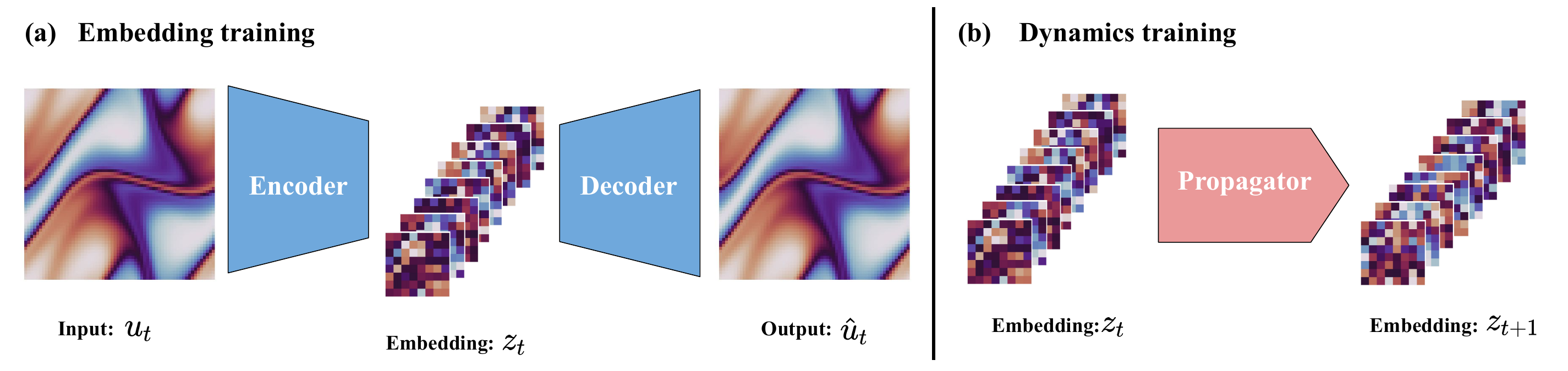}
    \caption{(a) An autoencoder is trained to project the input field to latent field with much coarser discretization. (b) A neural network is trained to predict the latent field at different time steps autoregressively.}
    \label{fig:schematic}
\end{figure}

\section{Methodology}

\subsection{Problem definition}

We are interested in solving time-dependent PDEs of the following form:
\begin{align}
     \frac{\partial u(\mathbf{x}, t)}{\partial t} &= F(u(\mathbf{x}, t), t, \theta),  \quad \mathbf{x} \in \Omega, t \in [0, T]\\
     u(\mathbf{x}, 0)& = u_0(\mathbf{x}), \quad \mathbf{x} \in \Omega,
\end{align} 
 where $T$ denotes the time horizon, $\Omega$ the spatial domain, $\theta$ are system parameters such as external forcing magnitude, and some boundary condition for $\mathbf{x} \in \partial \Omega$ is provided \textit{a priori} where $\partial \Omega$ denotes the boundary of the domain. To solve this initial value problem, we assume the system is Markovian such that $\frac{\partial u(\mathbf{x}, t)}{\partial t} = F(u(\mathbf{x}, t), \theta)$, parameters $\theta$ are time-invariant, and a neural network is trained to approximate the following time-discretized mapping:
 \begin{equation}
     u(\mathbf{x}, t+\Delta t)= \mathcal{A}(u(\mathbf{x}, t)),
 \end{equation}
 with a fixed $\Delta t$, and the system is assumed to be Markovian such that $u(\mathbf{x}, t+2\Delta t)=\mathcal{A}( \mathcal{A}(u(\mathbf{x}, t)))$.

 In practice, the function of interest at a particular time step $u(\cdot, t)$ is sampled on a $m$-point discretization grid $D$.  For a hierarchical model like U-Net, the grid will be altered internally between different layers and the mapping $\mathcal{A}$ is a composition of a sequence of mapping $\{\mathcal{A}_0, \hdots, \mathcal{A}_l\}$ which are approximated on grids $\{D_0, \hdots, D_l\}$ with $D_0=D$ and the number of grid points $m_l < m_{l-1} < \cdots < m_0$. In contrast to the aforementioned hierarchical model, we propose to learn $\mathcal{A}$ only on the coarsest grid $D_l$. 
 
\subsection{Autoencoder for discretized field's dimensionality reduction}

One of the most straightforward ways to project the function from the original grid to a coarser grid is through interpolation (\textit{e.g.}, bicubic interpolation). However, interpolation can result in significant information loss about the function, as a coarser grid can only evaluate a limited bandwidth and cannot distinguish frequencies that are higher than the Nyquist frequency. To achieve a less lossy compression of the input, we train a non-linear encoder network $\phi$ to project the input into latent space when coarsening its spatial grid. In the meantime, we train the decoder network $\psi$ to recover the input from the latent embedding that are represented on the coarse grid. The goal of training these two networks is to achieve data compression without too much loss of information such that their composition approximates an identity mapping: $I\approx\psi\circ\phi$ by minimizing the following lost function:
\begin{equation}
    L_{\text{AE}}= C\left(\psi(\phi(u)), u\right),
\end{equation}
where $C(\cdot, \cdot)$ is a distance function which we choose as $\mathcal{L}^2$ norm for all the systems. After the encoder and decoder network $\phi, \psi$ are trained, we fix their weights and train another propagator network $\gamma$ to predict the dynamics of the system: $z_{t+1}=\gamma(z_t), \text{where} \ z_t = \phi(u_t), z_{t+1}=\phi(u_{t+1})$. We empirically observe that training the autoencoder and dynamics propagator separately is crucial for final prediction accuracy (for more details we refer the reader to Section \ref{sec: ae ablation} in the supplementary information).

\begin{algorithm}[H]
\caption{First-stage training of encoder and decoder}\label{alg:ae-train}
\begin{algorithmic}[1]

\State \textbf{Input:} Training dataset $\{u^{(i)}_1, u^{(i)}_2, \hdots, u^{(i)}_T\}_{i=1}^N$, encoder network $\phi(\cdot)$, decoder network $\psi(\cdot)$, distance function $C(\cdot, \cdot)$, total optimization steps $K$.

\State Initialize $\phi, \psi$ randomly, set \textit{Iter}$=1$.
\For{\textit{Iter} $<K$}
   
    \State Sample a random snapshot $u_t^{(i)}$ from the dataset, where {\small$ t \sim \text{Uniform}(\{1, 2, \hdots, T\})$} 
    \State Reconstruct the data with autoencoder: $\hat{u}_t^{(i)} = \psi(\phi(u_t^{(i)}))$
    \State Compute reconstruction loss: $L_{\text{AE}} = C(\hat{u}_t^{(i)}, u_t^{(i)})$
    \State Take gradient descent step based on: $\nabla_\psi L_{\text{AE}}$, $\nabla_\phi L_{\text{AE}}$
    \State $\textit{Iter} \gets \textit{Iter} + 1$

\EndFor
\State \textbf{Output:} Trained encoder $\phi$ and decoder $\psi$.
\end{algorithmic}
\end{algorithm}
\vspace{-3mm}
\begin{algorithm}[H]
\caption{Second-stage training of the dynamics propagator}\label{alg:prop-train}
\begin{algorithmic}[1]

\State \textbf{Input:} Training dataset $\{u^{(i)}_1, u^{(i)}_2, \hdots, u^{(i)}_T\}_{i=1}^N$, trained encoder/decoder $\phi, \psi$, distance function $C(\cdot, \cdot)$, total optimization steps $K$, training rollout horizon $L$.

\State Encode the training dataset into a mesh-reduced latent space: $\{z^{(i)}_1, z^{(i)}_2, \hdots, z^{(i)}_T\}_{i=1}^N$, where $z^{(i)}_t = \phi(u^{(i)}_t)$.
Initialize the propagator network $\gamma$ randomly, set \textit{Iter}$=1$.
\For{\textit{Iter} $<K$}
   
    \State Sample a chunk of consecutive frames with temporal length $L+1$ from the dataset: $\{z^{(i)}_t, z^{(i)}_{t+1}, \hdots, z^{(i)}_{t+L}\}$, where {\small$ t \sim \text{Uniform}(\{1, 2, \hdots, T-L\})$}
    \State Initialize $s=t$, $\hat{z}_{s}^{(i)} \gets z_{s}^{(i)}$
    \While{$s<L$}
        \State Perform next step prediction: $\hat{z}_{s+1}^{(i)} = \gamma(\hat{z}_{s}^{(i)})$
        \State $s \gets s+1$
    \EndWhile
    \State Compute dynamics prediction loss: $L_{\text{DYN}} = C\left(\{\hat{z}_s^{(i)}\}_{s=t+1}^{t+L}, \{z_s^{(i)}\}_{s=t+1}^{t+L}\right)$
        \State Take gradient descent step on: $\nabla_\gamma L_{\text{DYN}}$
        \State $\textit{Iter} \gets \textit{Iter} + 1$
\EndFor
\State \textbf{Output:} Trained propagator $\gamma$.
\end{algorithmic}
\end{algorithm}
In this work, we exploit the uniform grid structure,  so that the majority of the autoencoder can be parameterized with convolutional neural networks (CNN) which have been shown to be effective for compressing imagery data \citep{oord2018neural, esser2021taming, rombach2022highresolution}. On top of the CNN modules, we also introduce self-attention layers \citep{Attention-NIPS-2017} near the bottleneck for better capturing long-range dependency in the field, which is beneficial for image synthesis tasks \citep{zhang2019selfattention, rombach2022highresolution, ho2020denoising}. Given the $i$-th input feature vector $u_i \in \mathbb{R}^{d_c}$ with channel size $d_c$, the (self-)attention can be defined as:

\begin{equation}
    u'_i
    = \sum_{j=1}^{m}\alpha_{ij}v_j, \quad \alpha_{ij} = 
    \frac{\exp{\left(q_i \cdot k_j / \sqrt{d_c} \right)}}
    {\sum_{s=1}^{m}\exp{\left(q_i \cdot k_s / \sqrt{d_c})\right)}},
    \label{eq:self-attention}
\end{equation}
where: $q_i=W_q u_i, k_i=W_k u_i, v_i=W_v u_i$ respectively, and $W_q, W_k, W_v \in \mathbb{R}^{d_c\times d_c}$ are learnable weights. Attention is also closely related to the learnable kernel integral \citep{kovachki2021neural}: 
\begin{equation}
    u_{l+1}(x)=\int_{\Omega} \kappa_{}(x, y)u_l(y) dy,
\end{equation}
where $u_{l+1}$ is the output function, $u_l$ is the input function with respect to $l\text{-th}$ layer, and $\kappa(\cdot)$ is the kernel parameterized by the dot product attention. 
The above learnable kernel integral is often used as the basic building block for learning mappings between function spaces. It can be parameterized with attention by using \eqref{eq:self-attention} to compute the $\kappa$ under some discretization of the domain, which have been studied in several prior works\citep{cao2021galerkin, guo2022transformer, kissas2022learning, li2023transformer, hao2023gnot, li2023scalable}.In the bottleneck, we use standard self-attention while for higher-resolution feature map we opt for axial factorized attention \citep{li2023scalable} which has superior computational efficiency on multi-dimensional problems.

\subsection{Baseline neural PDE solver}

\paragraph{Fourier neural operator} The spectral convolution layer is first proposed in the Fourier Neural Operator (FNO) \citep{li2020fourier} as a parameterization of the learnable kernel integral \citep{kovachki2021neural}. It applies a discrete Fourier transform to the input and then multiplies the $k$-lowest modes with learnable complex weights. Given input function $u_l$, the spectral convolution computes the kernel integral as follows:
\begin{equation}
        u_{l+1}(x)=\int_{\Omega} \kappa(x-y)u_l(y) dy =\sum_{\xi_1=0}^{\xi_1^{\max}} \hdots \sum_{\xi_n=0}^{\xi_n^{\max}} \mathbf{W}_{j} \mathbf{c}_{j} \mathbf{f}_{j}(x), 
        \label{eq:spectral-conv}
\end{equation}
where $\mathbf{f}_j$ is the $j$-th ($j:=(\xi_1, \xi_2, \hdots, \xi_n)$) Fourier basis function: $\mathbf{f}_j(x)=\exp{(2 i \pi \sum_d \frac{x_d \xi_d}{m_d})}, \xi_d \in \{0, 1, \hdots, m_d\}$ with $m_d$ being the resolution along the $d$-th dimension, $x_d$ being the coordinate for $d$-th dimension, and $\mathbf{c}_j=<u_l, \mathbf{f}_j>$ denotes the inner product between an input function and the Fourier series, $\mathbf{W} \in \mathbb{C}^{(\xi_1^{\max} \times \xi_2^{\max} \times \hdots \xi_n^{\max}) \times d_c \times d_c }$ is the learnable weight with frequency mode of the input at $i$-th dimension truncated up to $\xi_i^{\max}$.
Unlike the CNN layer, spectral convolution is able to capture multi-scale features that correspond to different frequencies within a single layer. It is also computationally efficient on a uniform grid as the $c_j$ can be computed via fast Fourier Transformation (FFT). In the original FNO, the update scheme at each layer is designed as follows:
\begin{equation}
    u_{l+1} = \sigma(K(u_l) + w(u_l)),
\end{equation}
where $w$ is a learnable linear transformation and $K$ is the spectral convolution as defined in \eqref{eq:spectral-conv}. Subsequent FNO \citep{tran2023factfno, pathak2022fourcastnet, guibas2021adaptive} works have shown that it is usually beneficial to use an update scheme similar to a standard Transformer with a skip connection \citep{Attention-NIPS-2017, He2016Residual}:
\begin{equation}
    \begin{aligned}
    u' &= K(\tilde{u_l}) + u_l, \\
    u_{l+1} &= \text{FFN}(\tilde{u'}) + u',
\end{aligned}
\end{equation}
where $\text{FFN}(\cdot)$ is a two-layer point-wise MLP (also known as a feed-forward network in Transformer literature),  $\tilde{\Diamond}$ denotes a feature map $\Diamond$ that is normalized, common choices of normalization include instance normalization \citep{ulyanov2017instance} and layer normalization \citep{ba2016layer}. We use instance normalization for $u_l$ and layer normalization for $u'$. Such architectural design is usually called a "mixer" in the computer vision literatures (e.g. Conv-Mixer \citep{trockman2022patches}, MLP-Mixer \citep{tolstikhin2021mlpmixer}), so we refer to this variant of FNO as FNO-Mixer in the rest of the paper.

\paragraph{UNet}

In the previous section we have introduced an autoencoder for projecting a discretized field to a coarse grid and learning the dynamics of the system on the coarsest grid. As an alternative to operating on the coarsest grid, we also investigate learning the dynamics on multiple discretization levels. This can be done by establishing a skip connection between every level in the encoder and decoder, from the input grid $D_0$ to the coarsest grid $D_l$, resulting in a standard U-Net architecture.

\subsection{Conditioning}

 Since we adopt a Markovian setting across all the experiments, there is no context for a model to infer dynamical property of the system such as the oscillation frequency of the external forcing. To modulate the prediction of neural networks based on the conditioning information, we investigate conditioning strategies that are often used in image-related tasks and introduced to PDE problems recently in \citet{gupta2023towards}. The system parameter $\theta \in \mathbb{R}$ (in our problem we have only a single scalar parameter but it is straightforward to extend this formulation to multi-variable parameter) are first projected to a high-dimensional embedding $\mathbf{q} \in \mathbb{R}^{d_f}$ using the sinusoidal projection from \citet{Attention-NIPS-2017} followed by a feed-forward network:
 \begin{equation}
 \begin{aligned}
     q' &= { \left[\cos(\theta), \sin(\theta), \cos(\frac{\theta}{10000^{2/d}}), \sin(\frac{\theta}{10000^{2/d}}),\hdots, \cos(\frac{\theta}{10000^{2(d-1)/d}}), \sin(\frac{\theta}{10000^{2(d-1)/d}})\right]}, \\
     q &=\text{FFN}(q'),
 \end{aligned}
 \end{equation}
 where $d$ is the a hyperparameter of hidden dimension and $d_f=2d$.
Below we will illustrate how to modulate different layers with the conditioning parameter $q$. 
\paragraph{Conditioning for a feed-forward network} We use a strategy similar to "AdaGN" proposed in \citet{nichol21improveddiff} to modulate the input to the feed-forward network. Given input $u_l$, we inject the conditioning information $q$ by doing an element-wise multiplication before processing it with a feed-forward network $\text{FFN}_l$:
\begin{equation}
\begin{aligned}
    u_l' &= \text{FFN}_{\text{cond}}(q) \odot u_l, \\
    u_{l+1} &= \text{FFN}_l(u_l') + u_l.
\end{aligned}
\end{equation}
    
\paragraph{Conditioning for a spectral convolution layer}
We adopt the frequency domain conditioning strategy proposed
in \citet{gupta2023towards}. The conditioning embedding $q$ first projects to a complex embedding vector $\hat{\mathbf{q}} \in \mathbb{C}^{(\xi_1^{\max} \times \xi_2^{\max} \times \hdots \xi_n^{\max})}$:
$
    \hat{\mathbf{q}} = qW^l_{\text{real}} + \iu qW^l_{\text{imag}},
$
where $W^l_{\text{real}}, W^l_{\text{imag}} \in \mathbb{R}^{c \times (\xi_1^{\max} \times \xi_2^{\max} \times \hdots \xi_n^{\max})}$ are learnable weights.
Then the conditioned version of \eqref{eq:spectral-conv} is defined as:
\begin{equation}
        u_{l+1}(x)=\sum_{\xi_1=0}^{\xi_1^{\max}} \hdots \sum_{\xi_n=0}^{\xi_n^{\max}} \hat{\mathbf{q}}_j \mathbf{W}_{j}  \mathbf{c}_{j} \mathbf{f}_{j}(x), \quad j:=(\xi_1, \xi_2, \hdots ,\xi_n).
        \label{eq:cond spectral-conv}
\end{equation}

\paragraph{Conditioning for a convolution block} For the conditioning of convolutional layers in the autoencoder and UNet, we following the conditioning strategy that is widely used in diffusion models \citep{ho2020denoising}, which amounts to adding the conditioning embedding to the input of convolution layer:
\begin{equation}
\begin{aligned}
  u_l'&=u_l+\text{FFN}_{\text{cond}}(q), \\
 u_{l+1} &= \text{Conv}_l(u_l') + u_l . 
\end{aligned}
\end{equation}.

\section{Experiments}

We test out the proposed model on three types of time-dependent PDE problems. In addition, we use the model to predict a system with a varying parameter (oscillation frequency) for a two-phase flow.

\subsection{Problems}

\paragraph{Navier-Stokes equation}
The 2D Navier-Stokes equation we consider is the 2D flow dataset proposed in \citet{li2020fourier}, which is based on 2D Navier-Stokes equation with a vorticity formulation.  The vorticity form reads as:
\begin{equation}
    \begin{aligned} \frac{\partial \omega (\mathbf{x},t)}{\partial t} + \mathbf{u}(\mathbf{x},t) \cdot \nabla \omega (\mathbf{x},t) &= \nu \nabla^2 \omega (\mathbf{x},t) + f(\mathbf{x}) , && \mathbf{x} \in (0,1)^2,  t \in (0,T], \\
\nabla \cdot \mathbf{u}(\mathbf{x},t) &= 0, &&  \mathbf{x} \in (0,1)^2, t \in [0,T], \\
\omega (\mathbf{x},0) &= \omega_0(\mathbf{x}), && \mathbf{x} \in (0,1)^2,
\label{eq:kmflow}   
\end{aligned}
\end{equation}
where $\omega$ denotes vorticity: $\omega:=\nabla \times u$, the initial condition $\omega_0$ is sampled from the Gaussian random field: $\mathcal{N}(0, 7^{3/2}(-\Delta + I)^{-2.5})$, the boundary condition is periodic: $u([x_1+1, x_2], t)=u[x_1, x_2], t), u([x_1, x_2+1], t)=u([x_1, x_2], t), \forall x_1, x_2 \in(0,1)$, the viscosity coefficient $\nu$ is $1e-4$ (dimensionless, corresponds to a Reynolds number of roughly $200$) and the forcing term is defined as: $f(\mathbf{x})=0.1(\sin 2\pi(x_1+x_2)+\cos2\pi(x_1+x_2))$. We are interested in learning to simulate the system (by predicting vorticity) from $t=5$ to $t=35$ seconds. The reference numerical simulation data is generated via the pseudo-spectral method. The dataset contains $1000$ trajectories where we use $900$ for training and $100$ for testing.

\paragraph{Shallow water equation}

We consider the 2D shallow water equation that is proposed in \citet{gupta2023towards}, which use the SpeedyWeather library\footnote{https://github.com/SpeedyWeather/SpeedyWeather.jl} to perform the numerical simulation. The equation is defined as:
\begin{equation}
\begin{aligned}
\frac{\partial \omega}{\partial t} + \nabla \cdot (\mathbf{u}(\omega + f)) &=
F_\omega + \nabla \times \mathbf{F}_\mathbf{u} + (-1)^{n+1}\nu\nabla^{2n}\omega, \\
\frac{\partial \mathcal{D}}{\partial t} - \nabla \times (\mathbf{u}(\omega + f)) &=
F_\mathcal{D} + \nabla \cdot \mathbf{F}_\mathbf{u}
-\nabla^2(\tfrac{1}{2}(u^2 + v^2) + g\eta) + (-1)^{n+1}\nu\nabla^{2n}\mathcal{D}, \\
\frac{\partial \eta}{\partial t} + \nabla \cdot (\mathbf{u}h) &= F_\eta,
\end{aligned}
\end{equation}
where $\omega$ is the vorticity, $g$ is the gravitational acceleration, $f$ is the Coriolis parameter, $\eta$ is the interface height (which can be interpreted as the pressure term), $h$ is dynamic layer thickness, $\mathcal{D}=\nabla \cdot \mathbf{u}$ is the velocity divergence, $F_\zeta, F_\mathcal{D}, F_\eta, \mathbf{F}_\mathbf{u} = (F_u,F_v)$ are different types of forcing terms (we refer reader to the official documentation of SpeedyWeather for more details). The dataset contains 1400 training trajectories and 1400 testing trajectories, each trajectory is of length 21 days (we sample each frame with an interval of $12$h).
\paragraph{Tank sloshing}

\begin{figure}[H]
    \centering
    \includegraphics[width=0.90\linewidth]{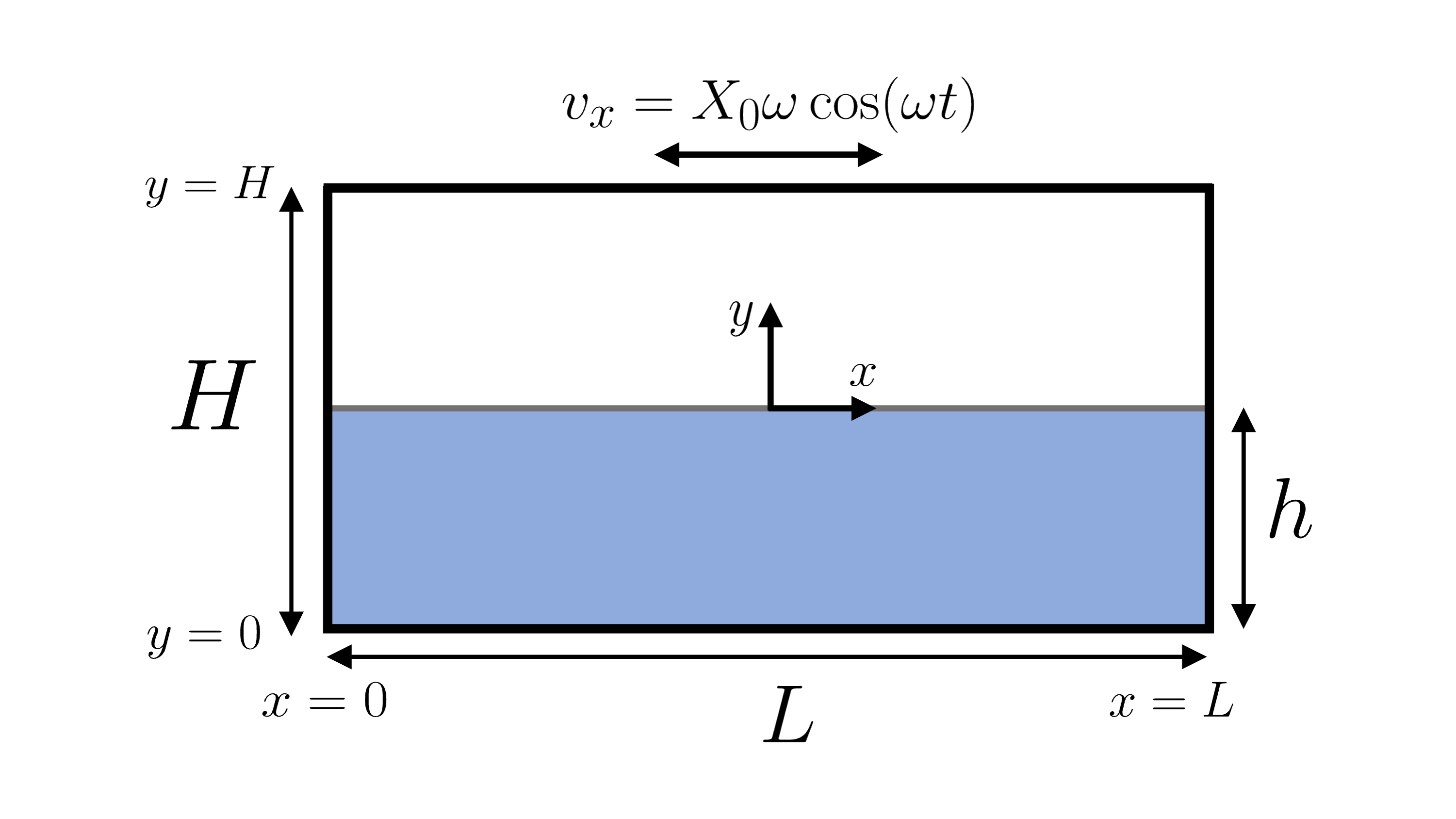}
    \caption{Description of the tank sloshing system}
    \label{fig:tank schematic}
\end{figure}
We consider the multi-phase problem of liquid sloshing in a partially filled 2D rectangular container subject to external perturbation. The motion of the liquid in the container is assumed to be incompressible, viscous, and turbulent, and we subject the container to a forced sinusoidal horizontal oscillation at a specific frequency $\omega$ and amplitude $X_0$. No-slip boundary conditions are applied at the boundaries of the container. Therefore, we use the Reynolds-Averaged Navier-Stokes (RANS) equations \cite{reynolds1895iv} to model the large scale mean fluid flow, and a turbulence closure model to describe the Reynolds stress induced by the velocity fluctuation terms. The Reynolds-Averaged continuity and Navier-Stokes equations are defined as:
\begin{equation}
\label{eqn:2pNSE}
    \begin{aligned}
         \nabla \cdot (\rho_m \bar{\mathbf{u}}) &= 0 \\
         \frac{\partial (\rho_m\bar{\mathbf{u}})}{\partial t} + \nabla \cdot (\rho_m \bar{\mathbf{u}}\bar{\mathbf{u}})&= - \nabla \bar{p} + \nabla \cdot [\mu_{e, m} (\nabla \bar{\mathbf{u}} + (\nabla \bar{\mathbf{u}})^T)  - \frac{2}{3} \rho_m \kappa \mathbf{I}]      + \rho_m \mathbf{g} + \rho_m \mathbf{f_s}\\
         \mu_{e, m} &= (1 - \alpha) \mu_{g}  + \alpha \mu_{l}  + \mu_{t} \\
         \rho_{m} &= (1-\alpha) \rho_g + \alpha \rho_l \\
\mathbf{\bar{u}}_{\Gamma_1} &= 0\\
\mathbf{\bar{u}}_{\Gamma_2} \cdot \mathbf{\hat{n}} &= 0
    \end{aligned}
\end{equation}
where $\alpha(\mathbf{x}, t)$ is the phase fraction of liquid, $\Gamma_1$ is the boundary of the container, $\mathbf{I}$ is the identity tensor. Additionally, $\mathbf{g}$ is the gravitational force, $\kappa$ is the turbulent kinetic energy, $\Gamma_2$ is the boundary between the liquid and gas phases, and $\rho_{m}$ is the mixture density accounting for the weighted contributions of the gas density, $\rho_g$ and the liquid density, $\rho_l$.  To model the turbulent stress, we use the two equation $k-\epsilon$ model to describe the conservation of turbulent kinetic energy, $k$, and the rate of dissipation of turbulent kinetic energy, $\epsilon$ \cite{launder1983numerical} (more details are provided in the supplementary information).  The effective mixture viscosity is given by $\mu_{e,m}$, which accounts for the contributions of the turbulent viscosity, $\mu_{t}$, and the weighted contributions of the gas viscosity $\mu_g$ and the liquid viscosity $\mu_{l}$.  $\mu_{t}$ is defined in terms of $k$ and $\epsilon$ according to Equation \ref{eq:turbulent_viscosity}.
The container motion is prescribed as an acceleration term, $\mathbf{f_{s}}$, where $\mathbf{f_s}$ is given by
\begin{equation}
\begin{aligned}
\mathbf{f_s} &= - X_0\omega^2 \sin(\omega t)\mathbf{e_{1}}\\
\end{aligned}
\end{equation}
 in which $X_0$ is the amplitude of the container motion, $\omega$ is the angular frequency of the container oscillation, $\rho$ is the density of the fluid within the container, and $\mathbf{e_1}$ is the horizontal unit vector.


The Volume of Fluid method is implemented to resolve the interface behavior between phases. To do so, $\alpha$ is defined to represent the volume fraction of liquid within each mesh element. The advection equation for $\alpha$ is defined as
\begin{equation}
\begin{aligned}
 \frac{\partial \alpha}{\partial t} + \nabla \cdot ( u \alpha )  = 0
\end{aligned}
\end{equation}

ANSYS \textsuperscript{®} FLUENT (v2023R2, ANSYS, Inc., Canonsburg, PA) is used to create datasets for two experiments. In the first experiment, 248 sloshing cases are created at angular frequencies ranging from $\omega$ = 1.0 rad/s to $\omega$ = 9.9 rad/s. The container is initialized at rest before the forced oscillation is applied for a duration of 8.0 s. Referring to (Figure \ref{fig:tank schematic}) and Equation \eqref{eqn:2pNSE}, we take $L = 1.2m$, $H=0.6m$, $\mu_l = 1\times 10^{-3}$ $\mathrm{Pa \cdot s}$, $\rho_l = 9.98 \times 10^{2} \: \mathrm{kg \: m^{-3}}$, $\mu_g = 1.8\times 10^{-5}$ $\mathrm{Pa \cdot s}$, and $\rho_g =1.23 \: \mathrm{kg \: m^{-3}}$. We let 40\% of the container volume be occupied by liquid and the remainder be occupied by gas. The intent is to model an air-water system. In the second experiment, the frequency is held constant at 3 rad/s and the initialized liquid level is varied from 10\% to 90\% of the total height of the container. These experiments are configured to avoid high-speed jet formation events caused by fluid impact with the container roof, and to span multiple distinct regimes of the sloshing behavior \cite{rognebakke2005sloshing, faltinsen2000multidimensional}.


\subsection{Implementation}
In this section, we list a brief summary of the experiment settings and for further details we refer reads to the Section \ref{sec: implementation} in the Appendix.
\paragraph{LNS Autoencoder} The encoder and decoder are mainly built upon convolutional layers. Internally they comprise a stack of downsampling/upsampling blocks, where each block downsamples/upsamples the spatial resolution by a factor of $2$. Each block contains a residual convolution block and a downsampling/upsampling layer. 
Self-attention layers are applied to the first two lowest resolution in the decoder part. Across all problems, the latent grid is $8$ times coarser than the original grid. For the $2$D Navier-Stokes problem, we set the latent resolution to $8\times8$ (original resolution: $64\times64$) and the latent dimension to $16$. For the two-phase flow problem, the latent resolution is $7\times15$ (original resolution: $61\times121$) and the latent dimension is $64$ . For the shallow-water equation, the latent resolution is $12\times24$ (original resolution: $96\times192$), with latent dimension also equals to $64$.

\paragraph{LNS Propagator} We use a simple residual network \citep{He2016Residual} to forecast the forward dynamics in the latent space, where each residual block contains three convolution layers with $3\times3$ convolution kernels and followed with a two-layer MLP. We employ dilated convolution for the middle convolution layer to capture multi-scale interaction. For the $2$D Navier-Stokes problem, we use $3$ residual blocks with a network width $128$. For other problems, we use $4$ residual blocks with a network width $128$ given the increased complexity of the problem.

\paragraph{Baselines} For FNO and FNO-Mixer, we use a layer of $4$ and hidden dimension of $64$, the modes used in different problems are marked as subscript in each result table, e.g. FNO$_{\xi_y,  \xi_x}$ means using $\xi_y$ modes in $y$-direction and $\xi_x$ modes in $x$-direction. For UNet in each problem, we use the same architecture as the autoencoder in LNS but use a skip connection at every level between the encoder and decoder.

\paragraph{Training} We first train the autoencoder by minimizing the relative $L^2$ reconstruction loss with a constant learning rate $3e-5$ using batch size of $32/64$ depending on the size of the system. We then train the propagator by minimizing the mean squared error between predicted embeddings and embed reference data for another 150k iterations with a learning rate of $5e-4$ and a cosine annealing schedule. For other models that operate on a full-order discretized space, we train it with a learning rate of $5e-4$ and a cosine annealing scheduling to minimize the relative $L^2$ prediction loss. During training, we also rollout the model to improve long-term stability. On the Navier-Stokes case, we rollout for $2$ steps during training and on tank sloshing problem, we rollout for $5$ steps during training. For these two systems, we find that further increasing the training rollout steps does not improve the testing performance significantly. On the shallow water equation, we observe that a longer rollout during training can greatly improve the the stability during training.

\subsection{Results}

In this section we will present the numerical results of different neural PDE solvers. We use relative $L^2$ norm between the predicted sequence $\hat{u}$ to the ground truth (GT) $u$ to measure the quantitative accuracy:
\begin{equation}
   \text{Error}=  \sqrt{\frac{\sum_{t=1}^T \sum_{i=1}^N (\hat{u}_{i, t}- u_{i, t})^2}{\sum_{t=1}^T \sum_{i=1}^N u_{i, t}^2}},
\end{equation}
with temporal discretization points $t \in {0, 1, \hdots, T}$ where $T$ denotes the temporal horizon, and spatial discretization points $i \in {0, 1, \hdots, N}$ where $N$ denotes the total number of spatial grid points. We also benchmark the computational cost of each model by profiling its training wall-clock time per iteration and peak GPU memory allocated.

\begin{figure}[H]
    \includegraphics[width=\linewidth]{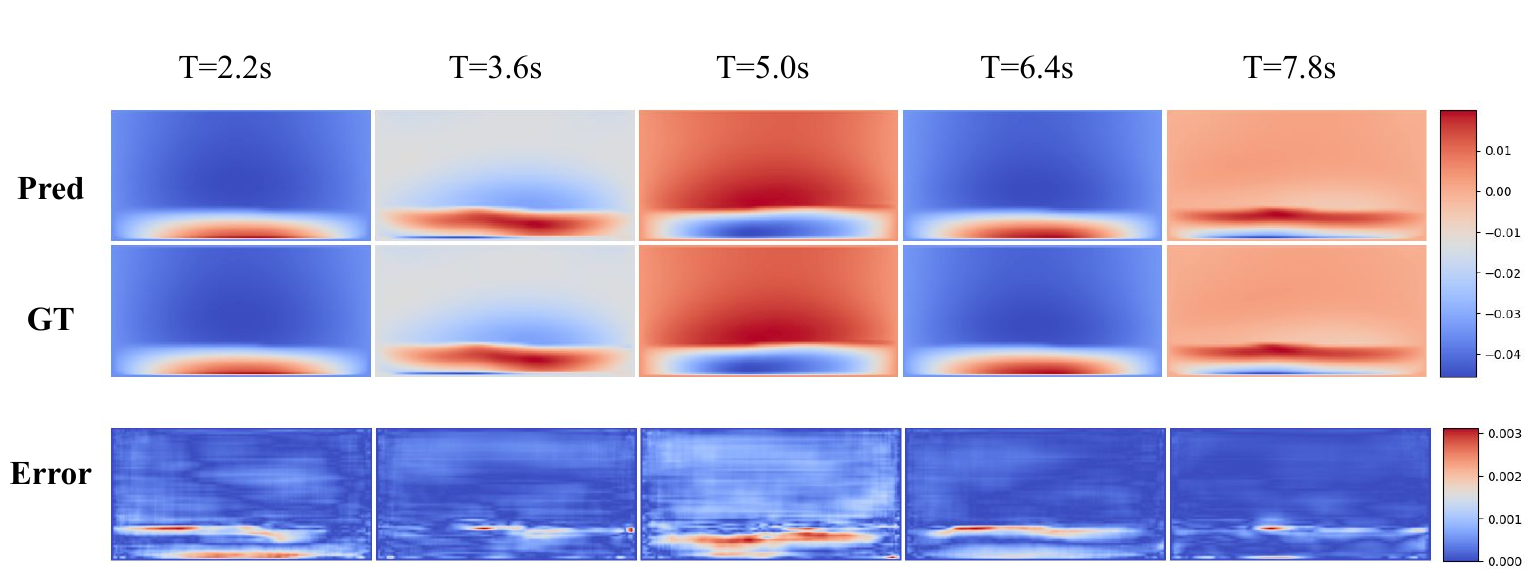}
    \caption{Tank Sloshing: Sample prediction of $x$ component of the velocity field $\mathbf{u}$ on \textit{varying height dataset} with liquid surface height: $h=25\%$. Unit: $m/s$.}
\label{fig:tank_slosh1}
\end{figure}

\begin{figure}[H]
    \includegraphics[width=\linewidth]{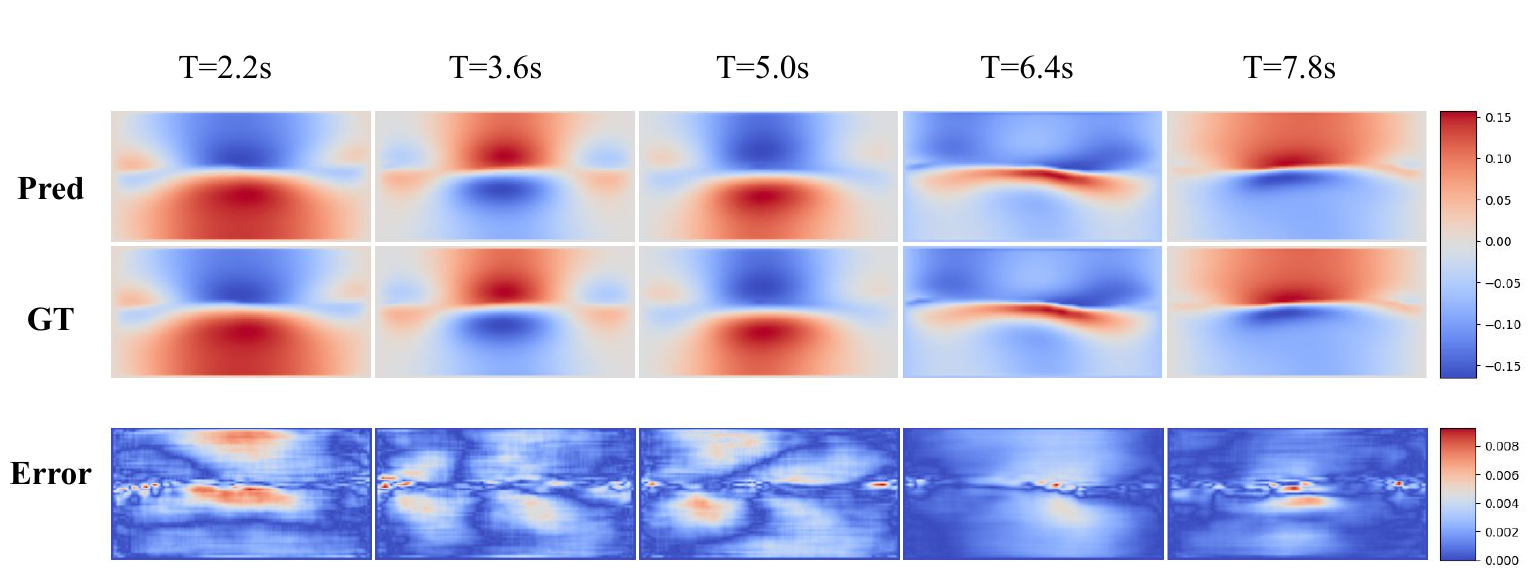}
    \caption{Tank Sloshing: Sample prediction $x$ component of the velocity field $\mathbf{u}$ on \textit{varying frequency dataset} with oscillation frequency: $\omega=7.12 ~\text{rad/s}$. Unit: $m/s$.}
\label{fig:tank_slosh2}
\end{figure}

As shown in Table \ref{tab:ns2d}, \ref{tab:tank-sloshing-height}, \ref{tab:tank-sloshing-freq} and Figure \ref{fig:tank_slosh1}, \ref{fig:tank_slosh2} we observe that all models can reach an error around or less than $15\%$ on the Navier-Stokes and tank sloshing problems. The best performing model varies case by case. FNO-Mixer shows a consistent improvement over FNO on these problems with increased computational cost. Notably, LNS achieves comparable results with UNet on these cases with greatly reduced computational cost. In addition, for the tank sloshing problem with varying frequency, all models can learn to predict the dynamics based on the conditioning parameter as demonstrated in Table \ref{tab:tank-sloshing-freq}.

\begin{figure}[H]
    \centering
    \includegraphics[width=\linewidth]{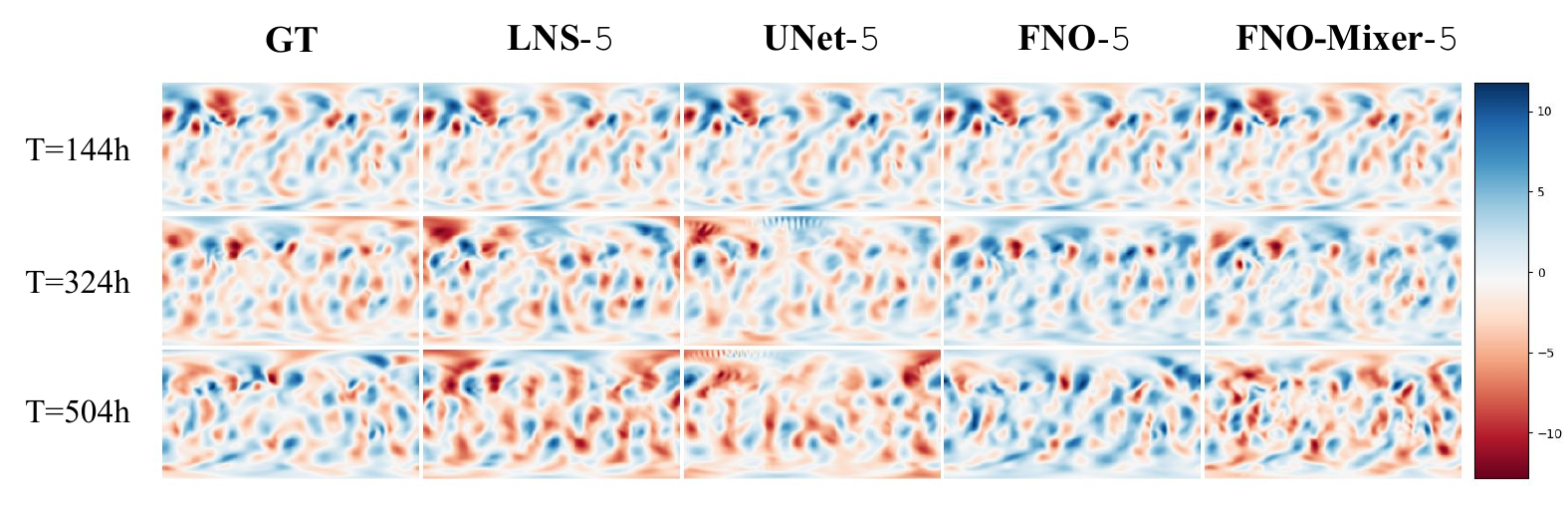}
    \vspace{-7mm}
    \caption{Shallow Water: Comparison on models' predicted $y-$component of $\mathbf{u}$ with $5$ training rollout steps. UNet's prediction exhibits notable artifacts after certain time steps. Unit: $m/s$.}
    \label{fig:sw-comparison-tw5}
\end{figure}

\begin{figure}[H]
    \centering
    \includegraphics[width=\linewidth]{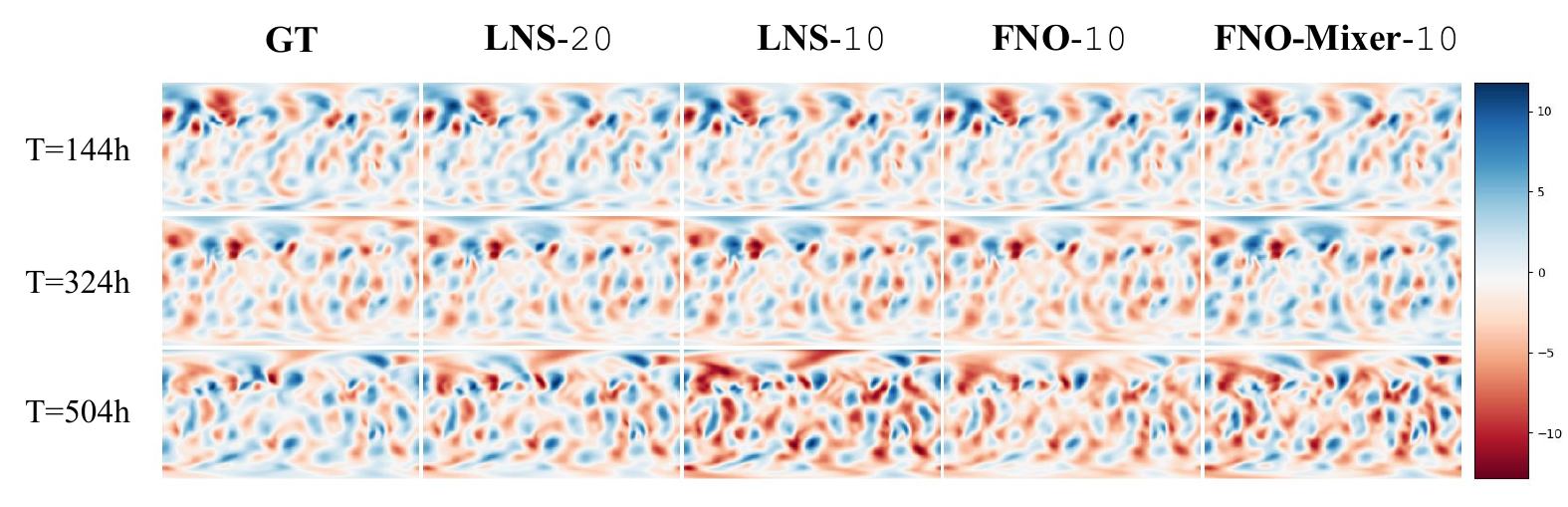}
    \vspace{-3mm}
    \caption{Shallow Water: Comparison on models' predicted $y-$component of $\mathbf{u}$ with longer training rollout.}
    \label{fig:sw-comparison-tw10}
\end{figure}

\begin{figure}[H]
    \centering
    \includegraphics[width=0.90\linewidth]{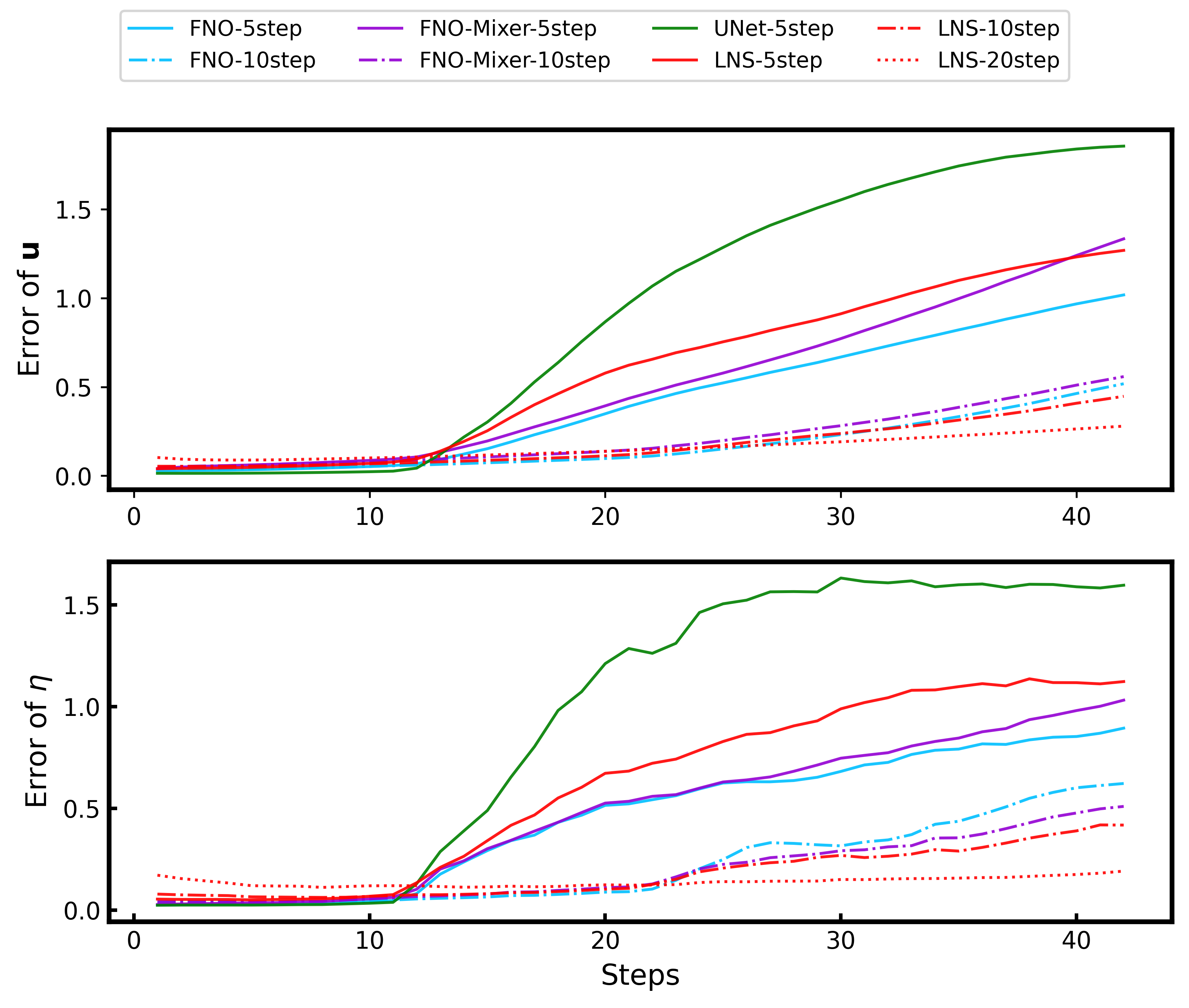}
    \vspace{-4mm}
    \caption{Comparison of rollout error trend of different models on the shallow water equation, the $\Delta t$ for each step is $12$h. The step number after each model's name indicates its training rollout steps.}
    \label{fig:sw rollout error trend}
\end{figure}

On more chaotic system - the shallow water equation, the performance discrepancy becomes more significant (see Table \ref{tab:sw_runtime} and Figure \ref{fig:sw-comparison-tw5}, \ref{fig:sw-comparison-tw10}). For UNet, despite its good prediction at the initial stage (green line in the Figure \ref{fig:sw rollout error trend})), it quickly blows up during the middle of the simulation and artifacts can be observed from the results. In general all the models are not able to maintain a stable long-term simulation (ending error greater than $100\%$ as shown in Figure \ref{fig:sw rollout error trend}) with only limited training rollout steps (5 steps). This blow-up error arises from the fact that these neural solvers do not have stability guarantee. The multi-step training target with backprop-through-time can effectively alleviate this. As given model's prediction $\hat{u}_{t+1}=\mathcal{A}_\theta(u_t)=u_{t+1}+\epsilon_t$ where $\epsilon_t$ is the prediction error of network at time step $t$, a multi-step rollout loss (e.g. using $L^2$ norm): $||\mathcal{A}_\theta(\mathcal{A}_\theta(\hdots\mathcal{A}_\theta(u_{t+1}+\epsilon_t)))-u_{t+L}||_2$ requires the model remain stable under perturbation $\epsilon_t$.  However, the computational cost increase linearly with respect to the rollout steps. Thanks to the highly compressed latent vector field in LNS, rolling out for more steps is much more affordable than models that operate on full-order space. Training LNS with training rollout steps up to $20$ is still highly efficient. As shown in the result, the longer training rollout steps greatly improves different models' stability. LNS trained with 20 rollout steps successfully maintain stable throughout the whole simulation and outperforms FNO and FNO-Mixer by a notable margin (Figure \ref{fig:sw rollout error trend}). It is worth noting that when training with more rollout steps, the model tends to trade-off next-step prediction accuracy for stability. At the beginning of simulation, LNS-5 and LNS-10 exhibit lower error compared to LNS-20. We hypothesize the major reason the data distribution in each training batches is affect by the rollout steps. The model with longer training rollout steps will observe more samples that are generated from its own prediction, whereas model trained with fewer rollout steps sees more samples from the dataset. At earlier time step during testing, the input will be much closer to the ground truth data distribution $p_{data}$ and gradually shift towards $p_{pred}$ due to error accumulation, thus the model that has trained with longer rollout tend to perform slightly worse at the beginning.

\begin{table}[h]
\centering
\begin{tabular}{cccc} 
\toprule
Model  & Rollout error & Time per iter (s)  & Peak Memory (Gb) \\ 
\midrule
$\text{FNO}_{16, 16}$    &   0.136            &   0.043      &    3.89    \\
$\text{FNO-Mixer}_{16, 16}$    &  \cellcolor{gray!12}0.061            &     \cellcolor{gray!12}0.055    &    \cellcolor{gray!12}4.91    \\

UNet   &     \cellcolor{gray!12}0.042          &       \cellcolor{gray!12}0.126      &   \cellcolor{gray!12}5.96 \\
LNS (Ours) &     0.074          &   0.064$+$0.017   &     4.43$+$2.97       \\
\bottomrule
\end{tabular}
\caption{(2D) Navier-Stokes equation with vorticity formulation. The computational runtime is benchmarked with a batch size of $32$ on a RTX3090 GPU. }
\label{tab:ns2d}
\end{table}

\begin{table}[H]
\centering
\begin{tabular}{cccccc} 
\toprule
\multirow{2}{*}{Model} & \multicolumn{3}{c}{Rollout error} & \multirow{2}{*}{Time per iter (s)} & \multirow{2}{*}{Peak Memory} \\
 & $\mathbf{u}$ & $p$ & $F$ &  &  \\ 
\midrule
$\text{FNO}_{16, 32}$ & 0.171 & 0.008 & 0.009 & 0.20 & 7.92  \\
$\text{FNO-Mixer}_{16, 32}$ & 0.119 &  0.006 & 0.008 & 0.25 & 13.32\\
UNet & \cellcolor{gray!12}0.101 & \cellcolor{gray!12}0.004 &  \cellcolor{gray!12}0.006 & \cellcolor{gray!12}0.33 & \cellcolor{gray!12}16.52 \\
LNS (Ours)& \cellcolor{gray!12}0.085 & \cellcolor{gray!12}0.006 & \cellcolor{gray!12}0.010 & \cellcolor{gray!12}0.054$+$0.029 & \cellcolor{gray!12}4.90$+$3.78  \\
\bottomrule
\end{tabular}
\caption{Tank sloshing - Varying height. The computational runtime is benchmarked with a batch size of $32$ on a RTX3090 GPU.}
\label{tab:tank-sloshing-height}
\end{table}
\begin{table}[H]
\centering
\begin{tabular}{cccccc} 
\toprule
\multirow{2}{*}{Model} & \multicolumn{3}{c}{Rollout error} & \multirow{2}{*}{Time per iter (s)} & \multirow{2}{*}{Peak Memory (Gb)} \\
 & $\mathbf{u}$ & $p$ & $F$ &  &  \\ 
\midrule
$\text{FNO}_{16, 32}$ & 0.042 & 0.002 & 0.004 & 0.21 & 9.29 \\
$\text{FNO-Mixer}_{16, 32}$ & \cellcolor{gray!12}0.030 & \cellcolor{gray!12}0.002  & \cellcolor{gray!12}0.003 & \cellcolor{gray!12}0.27 & \cellcolor{gray!12}14.38\\
UNet & \cellcolor{gray!12}0.037 & \cellcolor{gray!12}0.004 & \cellcolor{gray!12}0.002 & \cellcolor{gray!12}0.34 & \cellcolor{gray!12}16.94 \\
LNS (Ours)& 0.041 & 0.006 & 0.004 & 0.054$+$0.040 & 4.90$+$4.18 \\
\bottomrule
\end{tabular}
\caption{Tank sloshing - Varying frequency (Conditional prediction). The computational runtime is benchmarked with a batch size of $32$ on a RTX3090 GPU.}
\label{tab:tank-sloshing-freq}
\end{table}

\begin{table}[H]
\centering
\vspace{-2mm}
\begin{tabular}{cccccccc} 
\toprule
\multirow{2}{*}{Model} & Training & \multicolumn{2}{c}{Rollout error} & \multicolumn{2}{c}{\multirow{2}{*}{Time per iter (s)}} & \multicolumn{2}{c}{\multirow{2}{*}{Peak Memory (Gb)}} \\
 & Rollout steps & $\mathbf{u}$ & $\eta$ & \multicolumn{2}{c}{} & \multicolumn{2}{c}{} \\ 
\midrule
\multirow{3}{*}{$\text{FNO}_{16, 32}$} & 5 & 0.527 & 0.558 & \multicolumn{2}{c}{0.42} & \multicolumn{2}{c}{14.58} \\
 & 10 &  0.214 & 0.307 & \multicolumn{2}{c}{0.87} & \multicolumn{2}{c}{26.98} \\
 & 20 & - & - & \multicolumn{2}{c}{-} & \multicolumn{2}{c}{-} \\ 
\midrule
\multirow{3}{*}{$\text{FNO-Mixer}_{16, 32}$} & 5 & 0.668 & 0.656 & \multicolumn{2}{c}{0.54} & \multicolumn{2}{c}{25.10} \\
 & 10 & 0.271 & 0.336 & \multicolumn{2}{c}{1.09} & \multicolumn{2}{c}{46.19} \\
 & 20 & - & - & \multicolumn{2}{c}{-} & \multicolumn{2}{c}{-} \\ 
\midrule
\multirow{3}{*}{UNet} & 5 & 0.931 & 1.180 & \multicolumn{2}{c}{1.13} & \multicolumn{2}{c}{36.87} \\
 & 10 & - & - & \multicolumn{2}{c}{-} & \multicolumn{2}{c}{-} \\
 & 20 & - & - & \multicolumn{2}{c}{-} & \multicolumn{2}{c}{-} \\ 
\midrule
\multirow{3}{*}{LNS (Ours)} & 5 & 0.736 & 0.758 & \multirow{3}{*}{0.33$+$}& 0.10 & \multirow{3}{*}{17.39$+$} & 4.05 \\
 & \cellcolor{gray!12}10 & \cellcolor{gray!12}0.198 &\cellcolor{gray!12}0.199 &  &\cellcolor{gray!12}0.19  &  & \cellcolor{gray!12}4.76 \\
 & \cellcolor{gray!12}20 & \cellcolor{gray!12}0.132 & \cellcolor{gray!12}0.113 &  & \cellcolor{gray!12}0.36 &  & \cellcolor{gray!12}6.42 \\
 \bottomrule
\end{tabular}
\vspace{-3mm}
\caption{Shallow water equation. The computational runtime is benchmarked with a batch size of $32$ on a A6000 GPU. "-" indicates this training setting is too computationally expensive to carry out.}
\label{tab:sw_runtime}
\end{table}

\section{Conclusion}

In this work, we propose a new data-driven framework for predicting time-dependent PDEs and investigate different kinds of neural PDE solvers. We show that training the temporal model in the mesh-reduced space improves the computational efficiency and is beneficial for problems that feature latent dynamics distributed on a low-dimensional manifold. It also opens up the possibility for training model with longer backprop-through-time horizon, which can significantly improve models' stability. The observation in this study is also in alignment with the recent success of a series of image synthesis models that learn the generative model in the latent space instead of pixel space \citep{rombach2022highresolution,esser2021taming, vahdat2021scorebased}. As this work only considers uniform meshes, an interesting future direction would be the extension to arbitrary meshes and geometries.

\bibliography{ref}

\newpage
\begin{suppinfo}
\section{Details for model implementation}
\label{sec: implementation}
In this section we provide the detail for the model architecture and hyperparameter choice of different models considered in this paper.

\begin{figure}[h]
\vspace{-4mm}
\includegraphics[width=\linewidth]{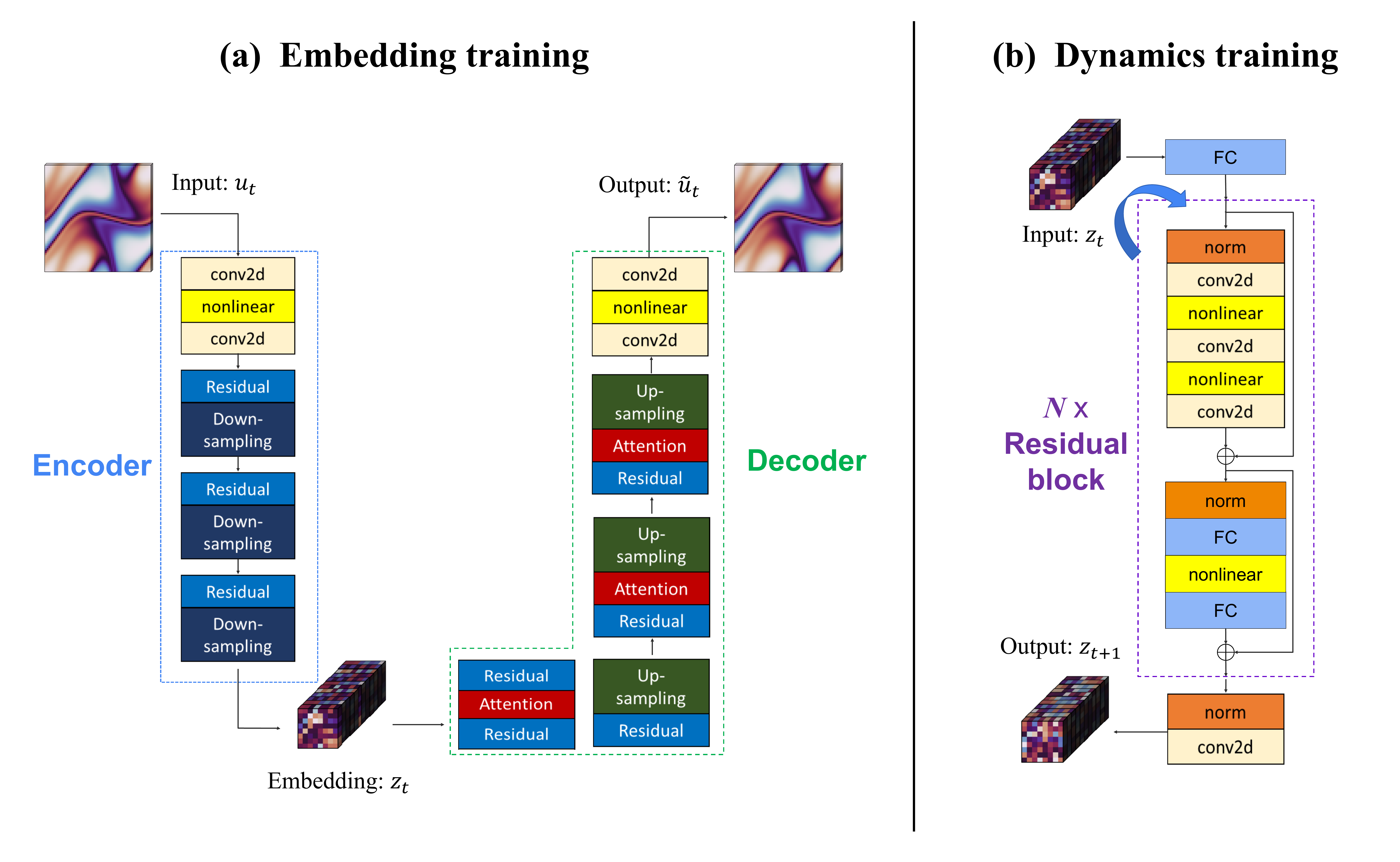}
\caption{
Detail architecture of the convolution based autoencoder and dynamics propagator. \textbf{"conv2d"}: a convolution layer with $3\times3$ kernel, stride 1 and padding 1; \textbf{"nonlinear"}: a non-linear activation function (we use Gaussian Error Linear Units (GELU) \citep{hendrycks2016gaussian} throughout different experiments
); \textbf{"FC"}: a fully-connected layer which applies learnable linear transformation to the input; \textbf{"Attention"}: self-attention layer defined in \eqref{eq:self-attention}; \textbf{"norm"}: normalization layer, where we use group normalization \citep{wu2018group} throughout different experiments. \label{fig:detail architecture}
}
\vspace{-4mm}
\end{figure}
The residual block is defined as a stack of convolutional layers and non-linear activation function with skip connection: $a^{(l+1)}=\text{conv2d} \circ\text{nonlinear}\circ\text{conv2d}(a^{(l)}) + a^{(l)}$. For the downsampling layer, it is a convolution layer with $3\times3$ kernel and a stride of $2$, no padding. This downsamples the mesh resolution by $2$ along each axis. For the upsampling, it is a composition of nearest interpolation and a convolution layer with $3\times3$ kernel, stride $1$ and padding $1$. 

\newpage
\begin{table}[H]
    \begin{subtable}{.4\linewidth}
      \centering
        \caption{FNO Hyperparameters}
        \begin{tabular}{lc}
            \toprule
              Parameter & Setting \\
                        
            \cmidrule(rl){1-2}
            
              Modes & 16 \\
              Width & 64 \\
              \# Blocks & 4 \\
              Init lr & 5e-4\\
             Lr schedule & Cosine Aneal \\
             Training Epochs & 500 \\
             Batch size & 32 \\
            \bottomrule
        \end{tabular}
    \end{subtable}%
    \begin{subtable}{.5\linewidth}
      \centering
        \caption{UNet Hyperparameters}
        \begin{tabular}{lc}
            \toprule
            Parameter & Setting\\
                        
            \cmidrule(rl){1-2}
            
              Width & (64, 64, 128, 128) \\
              Attn & (false, false, false, true) \\
              \# res. blocks & 2 \\
              Init lr & 5e-4\\
              Lr schedule & Cosine Aneal \\
               Training Epochs & 500 \\
             Batch size & 32 \\
             \bottomrule
        \end{tabular}
    \end{subtable} 
    
    \vspace{+3mm}
    
    \begin{subtable}{.4\linewidth}
      \centering
        \caption{(Latent) Resnet Hyperparameters}
        \begin{tabular}{lr}
            \toprule
            Parameter & 1D/2D\\
                        
            \cmidrule(rl){1-2}
            
              Width & 128 \\
             Dilation & 2 \\
              \# blocks & 3 \\
              Init lr & 5e-4 \\
              Lr schedule & Cosine Aneal \\
             Training Epochs & 500 \\
             Batch size & 32 \\
             \bottomrule
        \end{tabular}
    \end{subtable} 
   \begin{subtable}{.5\linewidth}
      \centering
        \caption{Autoencoder Hyperparameters}
        \begin{tabular}{lc}
            \toprule
            Parameter & Setting\\
                        
            \cmidrule(rl){1-2}
            
              Width & (64, 64, 128, 128) \\
              Attn & (false, false, false, true) \\
              \# res. blocks & 2 \\
              Latent resolution & $8 \times 8$\\
              Latent channel &  $16$ \\
              Init lr & 3e-5\\
              Lr schedule & Constant \\
              Training Epochs & 500 \\
             Batch size & 32 \\
             \bottomrule
        \end{tabular}
    \end{subtable} 
    \caption{Hyperparameters for architectures used for 2D Navier-Stokes. For convolution layers that involve padding, we use circular padding in all directions.  For the specification of UNet and Autoencoder, the width indicates the number of feature channels at different resolutions (the first one corresponds to the input grid and the last one corresponds to the latent grid), and Attn (attention) indicates if any attention mechanism is used at the specific resolution. \# res. blocks indicates how many residual blocks are used at each resolution.}

    \label{tab:model params ns2d}
        \vspace{-2mm}
\end{table}

\begin{table}[H]
    
    \begin{subtable}{.4\linewidth}
      \centering
        \caption{FNO Hyperparameters}
        \begin{tabular}{lc}
            \toprule
              Parameter & Setting \\
                        
            \cmidrule(rl){1-2}
            
              Modes (y/x) & 32/16 \\
              Width & 64 \\
              \# blocks & 4 \\
              Init lr & 5e-4\\
             Lr schedule & Cosine Aneal \\
              Training Epochs & 500 \\
             Batch size & 32 \\
            \bottomrule
        \end{tabular}
    \end{subtable}%
    \begin{subtable}{.5\linewidth}
      \centering
        \caption{UNet Hyperparameters}
        \begin{tabular}{lc}
            \toprule
            Parameter & Setting\\
                        
            \cmidrule(rl){1-2}
            
              Width & (64, 64, 128, 128) \\
              Attn & (false, false, false, true) \\
              \# res. blocks & 2 \\
              Init lr & 5e-4\\
              Lr schedule & Cosine Aneal \\
               Training Epochs & 500 \\
             Batch size & 32 \\
             \bottomrule
        \end{tabular}
    \end{subtable} 
    
    \vspace{+3mm}
    
    \begin{subtable}{.4\linewidth}
      \centering
        \caption{(Latent) Resnet Hyperparameters}
        \begin{tabular}{lc}
            \toprule
            Parameter & Setting\\
                        
            \cmidrule(rl){1-2}
            
              Width & 128 \\
              Dilation & 3 \\
              \# blocks & 4 \\
              Init lr & 5e-4 \\
              Lr schedule & Cosine Aneal \\
            Training Epochs & 500 \\
             Batch size & 32 \\
             \bottomrule
        \end{tabular}
    \end{subtable} 
   \begin{subtable}{.5\linewidth}
      \centering
        \caption{Autoencoder Hyperparameters}
        \begin{tabular}{lc}
            \toprule
            Parameter & Setting\\
                        
            \cmidrule(rl){1-2}
            
              Width & (64, 64, 128, 128) \\
              Attn & (false, false, false, true) \\
              \# res. blocks & 2 \\
              Latent resolution & $12 \times 24$\\
              Latent channel &  $64$ \\
              Init lr & 3e-5\\
              Lr schedule & Constant \\
            Training Epochs & 500 \\
             Batch size & 32 \\
             \bottomrule
        \end{tabular}
    \end{subtable} 
    \caption{Hyperparameters for architectures used for 2D Shallow-water equation. For convolution layers that involve padding, we use circular padding in x-direction and constant padding (zero) in y-direction. }
    \label{tab:model params sw2d}
    \vspace{-2mm}
\end{table}

\begin{table}[H]
    
    \begin{subtable}{.4\linewidth}
      \centering
        \caption{FNO Hyperparameters}
        \begin{tabular}{lc}
            \toprule
              Parameter & Setting \\
                        
            \cmidrule(rl){1-2}
            
              Modes (y/x) & 32/16 \\
              Width & 64 \\
              \# blocks & 4 \\
              Init lr & 5e-4\\
             Lr schedule & Cosine Aneal \\
            Training Epochs & 100 (80) \\
             Batch size & 32 (16) \\
            \bottomrule
        \end{tabular}
    \end{subtable}%
    \begin{subtable}{.5\linewidth}
      \centering
        \caption{UNet Hyperparameters}
        \begin{tabular}{lc}
            \toprule
            Parameter & Setting\\
                        
            \cmidrule(rl){1-2}
            
              Width & (64, 64, 128, 128) \\
              Attn & (false, false, false, true) \\
              \# res. blocks & 2 \\
              Latent resolution & $7 \times 15$\\
              Latent channel &  $64$ \\
              Init lr & 5e-4\\
              Lr schedule & Cosine Aneal \\
              Training Epochs & 100 \\
             Batch size & 32 \\
             \bottomrule
        \end{tabular}
    \end{subtable} 
    
    \vspace{+3mm}
    
    \begin{subtable}{.4\linewidth}
      \centering
        \caption{(Latent) Resnet Hyperparameters}
        \begin{tabular}{lc}
            \toprule
            Parameter & Setting\\
                        
            \cmidrule(rl){1-2}
            
              Width & 128 \\
              Dilation & 3 \\
              \# blocks & 4 \\
              Init lr & 5e-4 \\
              Lr schedule & Cosine Aneal \\
            Training Epochs & 100 \\
             Batch size & 32 \\
             \bottomrule
        \end{tabular}
    \end{subtable} 
   \begin{subtable}{.5\linewidth}
      \centering
        \caption{Autoencoder Hyperparameters}
        \begin{tabular}{lc}
            \toprule
            Parameter & Setting\\
                        
            \cmidrule(rl){1-2}
            
              Width & (64, 64, 128, 128) \\
              Attn & (false, false, false, true) \\
              \# res. blocks & 2 \\
              Latent resolution & $7 \times 15$\\
              Latent channel &  $64$ \\
              Init lr & 3e-5\\
              Lr schedule & Constant \\
               Training Epochs & 100 \\
             Batch size & 32 \\
             \bottomrule
        \end{tabular}
    \end{subtable} 
    \caption{Hyperparameters for architectures used for Tank sloshing problem. For convolution layers that involve padding, we use constant padding. When using longer training rollout horizon in FNO, we decrease the batch size to reduce computational cost. }
    \label{tab:model params two phase}
    \vspace{-2mm}
\end{table}
\newpage
\section{Details for tank sloshing data generation}

The equations that define the two equation $k-\epsilon$ model are reproduced below:


\begin{equation}
\begin{aligned}
\frac{\mathrm{D} (\rho k)}{\mathrm{D}t} &= \frac{\partial}{\partial x_j} \left[ \left (\mu + \frac{\mu_t}{\sigma_k} \right ) \frac{\partial k}{\partial x_j} \right] + \mu_t \left( \frac{\partial \bar{u_i}}{\partial x_j} \frac{\partial \bar{u_i}}{\partial x_j} + \frac{\partial \bar{u_j}}{\partial x_i} \frac{\partial \bar{u_i}}{\partial x_j} \right) - \rho \epsilon, \\
\frac{\mathrm{D} (\rho \epsilon)}{\mathrm{D}t} &= \frac{\partial}{\partial x_j} \left[\left ( \mu + \frac{\mu_t}{\sigma_\epsilon} \right ) \frac{\partial \epsilon}{\partial x_j} \right] + C_1 \frac{\epsilon}{k} \mu_t \left( \frac{\partial \bar{u_i}}{\partial x_j} \frac{\partial \bar{u_i}}{\partial x_j} + \frac{\partial \bar{u_j}}{\partial x_i} \frac{\partial \bar{u_i}}{\partial x_j} \right) - C_2 \frac{\rho \epsilon^2}{k}
\end{aligned}
\end{equation}
where $\sigma_k$ and $\sigma_{\epsilon}$ are the Prandtl numbers relating the diffusion of turbulent kinetic energy $k$ and the rate of turbulent kinetic energy dissipation $\epsilon$ to the turbulent viscosity $\mu_t$ ($\sigma_k = 1.0$, $\sigma_{\epsilon} = 1.3$). $C_1$, $C_2$ are empirical constants ($C_1 = 1.44$, $C_2 = 1.92$). The turbulent viscosity is defined based on $k$, $\epsilon$, and the constant $C_\mu$ as

\begin{equation}
    \begin{aligned}
    \label{eq:turbulent_viscosity}
        \mu_{t} &= \rho C_{\mu} \frac{k^2}{\epsilon}
    \end{aligned}
\end{equation}
where $C_{\mu} = 0.09$.

The tank sloshing simulation is solved using the Pressure Implicit Splitting of Operators (PISO) scheme, which decomposes the solution operators into implicit prediction and explicit correction steps \citep{issa1986solution}. The  PRESTO (Pressure Staggering) pressure correction scheme is used to interpolate between the cell-defined pressure values to determine the pressure defined at the mesh faces, which is necessary for the momentum update equations \citep{patankar1982numerical}.  

The simulation is conducted using a uniform grid mesh, where the size of each element is $\Delta x = 0.01 \: \mathrm{m}$, for a total of 7200 grid cells. A uniform timestep of $\Delta t = 0.005$ s is used with a first-order implicit timestepping scheme. A sensitivity study was performed to identify a suitable spatio-temporal discretization able to resolve the dynamics of the process (Figure \ref{fig:spatial, time convergence}). The transient variation of the gauge pressure at a reference point $x = 0.6 \: \mathrm{m},\; y = 0.36 \: \mathrm{m}$ within the domain is examined as a metric for simulation mesh-independence, for mesh elements 0.25$\times$, 0.5$\times$,  2$\times$, and 4$\times$ the size of the original mesh grid size. A similar study is used to identify a suitable time discretization, with the transient gauge pressure examined at 0.25$\times$, 4$\times$ and 8$\times$ the reference $\Delta t$ used for simulation. 
\begin{figure}[H]
    \centering
    \includegraphics[width=\linewidth]{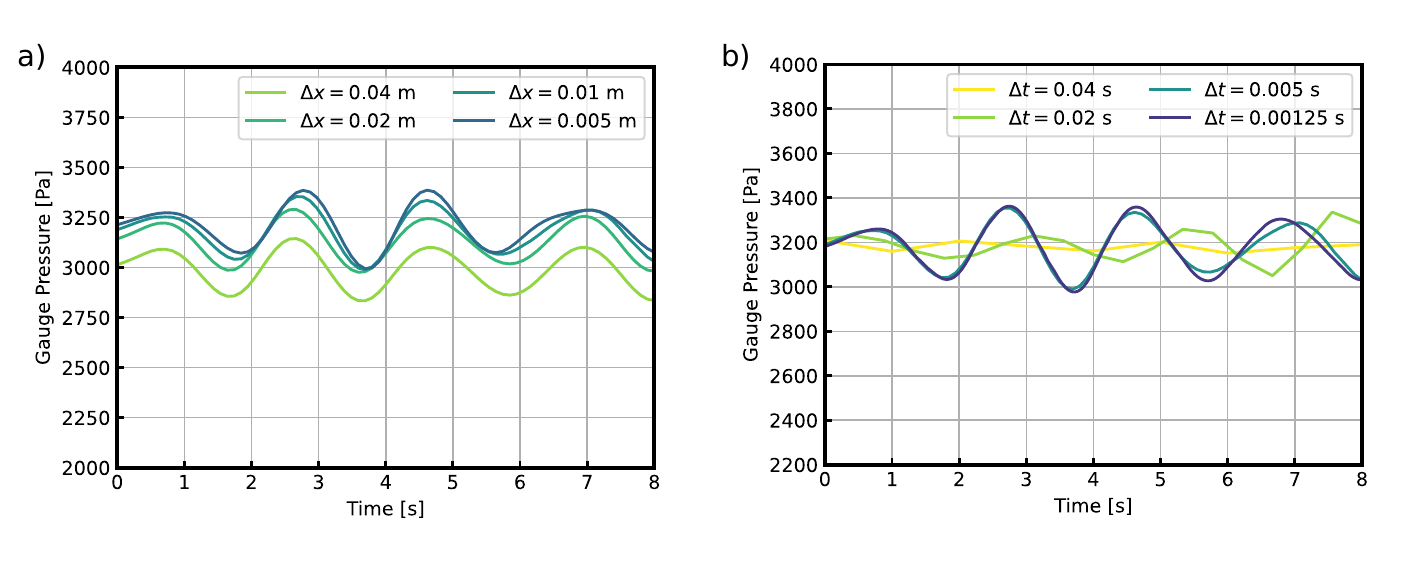}
    \vspace{-3mm}
    \caption{(a) A mesh convergence study on the tank sloshing simulation. The gauge pressure is taken at the reference point $x = 0.6$ m, $y = 0.36$ m, and the time increment is $\Delta t = 0.005$ s. The spatial discretization used in this work is $\Delta x$ = 0.01 m. (b) A time discretization convergence study on the tank sloshing simulation. The gauge pressure is taken at the reference point $x = 0.6$ m, $y = 0.36$ m. The spatial discretization is held constant at $\Delta x$ = 0.01 m, and the time increment of $\Delta t$ = 0.005 s is used in this work.}
    \label{fig:spatial, time convergence}
\end{figure}


\section{Ablation on encoder/decoder training} \label{sec: ae ablation}

In this section we present a study on the influence of different training strategies of the encoder/decoder ($\phi(\cdot), \psi(\cdot)$) and the propagator network ($\gamma(\cdot)$). The first strategy is to train the propagator network and encoder/decoder separately as we have introduced in the main text, which we will refer to as "Two stage" training in the below discussion. The training of the encoder/decoder can also be combined with the propagator, where the networks' parameters are optimized by minimizing the combination of reconstruction loss and dynamics prediction loss (the training rollout steps is $m$):
\begin{equation}
    L = \frac{1}{L}\sum_{m=1}^{m=L}||u_{t+m} - \psi(\underbrace{\gamma \circ \gamma \circ \hdots \circ \gamma}_{\times m}\left(\phi(u_t)\right))||_2^2.
\end{equation}
In addition, we can combine the encoder/decoder with the propagator network together $\gamma_{\text{combined}}=\psi \circ \gamma \circ \phi$, which is equivalent to a UNet without any skip connections but only a bottleneck. We denote the model trained on the combined loss as "Combined loss" and the model formed by combining three sub-networks as "Combined architecture". Lastly, we study another popular encoding scheme where the discretization will not be altered during encoding/decoding process - the encode-process-decode (EPD) scheme \citep{stachenfeld2022learned, pfaff2021learning, sanchez2020learning}. More specifically, we compared against the Dilated CNN with MLP-based encoder/decoder used in \citet{stachenfeld2022learned}.
\begin{figure}
    \centering
    \includegraphics[width=0.70\linewidth]{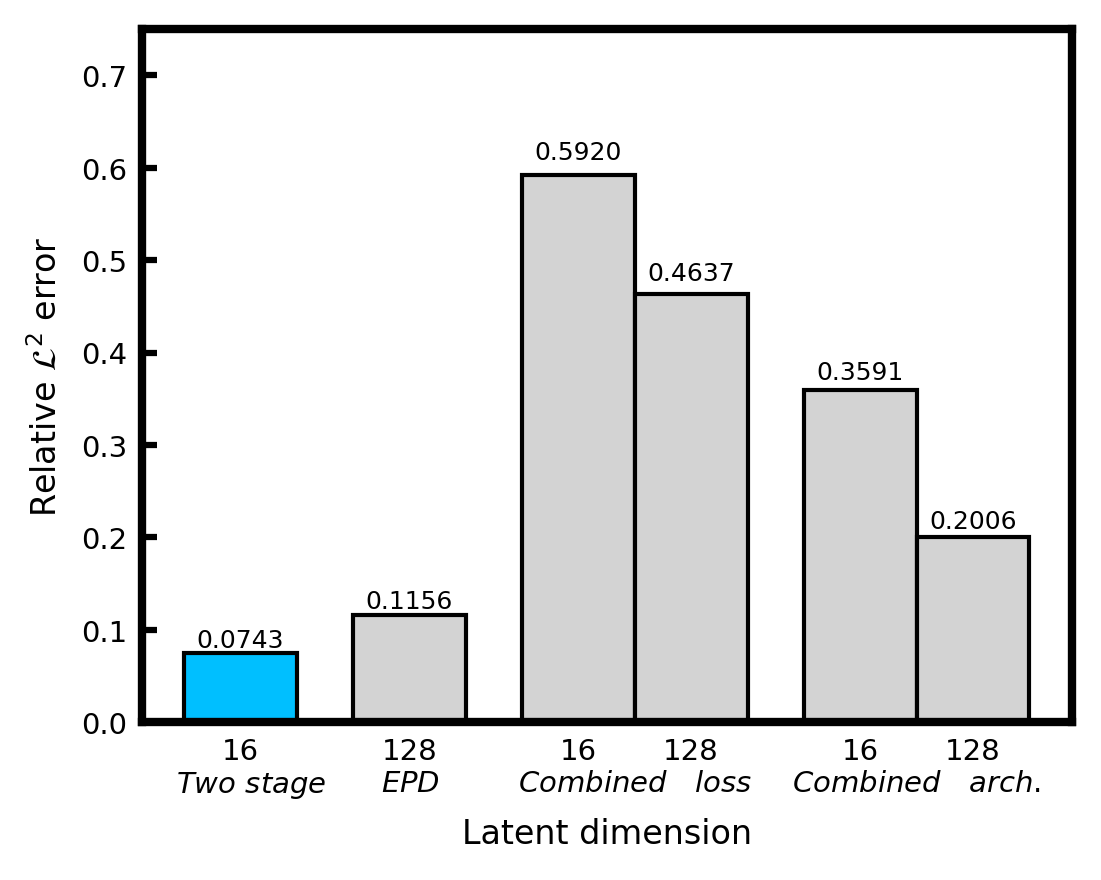}
    \caption{Rollout error of different models on 2D Navier-Stokes equation.}
    \label{fig:training ablation}
\end{figure}

As shown in Figure \ref{fig:training ablation}, training encoder/decoder jointly with propagator results in a degraded performance especially if the dimension of the bottleneck are chosen to be relatively small. With a larger latent dimension, the jointly trained model still cannot match the performance of a model that employs two stage training.  We hypothesize the degradation of the performance of joint training is due to the difficulty of optimization. When jointly training the autoencoder and the propagator, the encoder is being updated at every optimization step and thus the latent dynamics for propagator to approximate is always changing along the course of training. This poses an optimization challenge for the propagator network. On the other hand, the encoder/decoder need to balance between learning an embedding space that's easier for propagator to predict the forward dynamics and an embedding space that can achieve minimum reconstruction error. When training the autoencoder and dynamics propagator separately, the training objective is de-coupled so it eases the difficulty of the optimization process without the need to balance the reconstruction and dynamics propagation loss with a carefully selected loss weight. For the combined architecture variant, it under-performs two stage model and also the UNet benchmarked in the main manuscript. Compared to UNet, it does not have multi-scale skip connection. This increases the difficulty for the optimization as mesh reduction at each level will lose some high-frequency information, while in UNet the skip-connection at different scales can preserve these high-frequency features. It is worth noting that in two stage training the encoder/decoder also does not have skip connection to preserve the high-frequency features on the higher resolution grid, but it is trained to learn to reconstruct them in the first stage training. We postulate that reconstructing high-frequency components is easier than predicting high-frequency components of the future frames, which results in a combined architecture model variant performs worse and requires much more latent channels to reach reasonable accuracy.
Furthermore, our proposed model is able to outperform the encode-process-decode scheme on this NS2D problem despite the dynamics propagator operates on a very coarse latent field.

\section{Visualization}

\subsection{Navier-Stokes 2D}

\begin{figure}[H]
\begin{subfigure}{\textwidth}
\centering
    \includegraphics[width=0.9\linewidth]{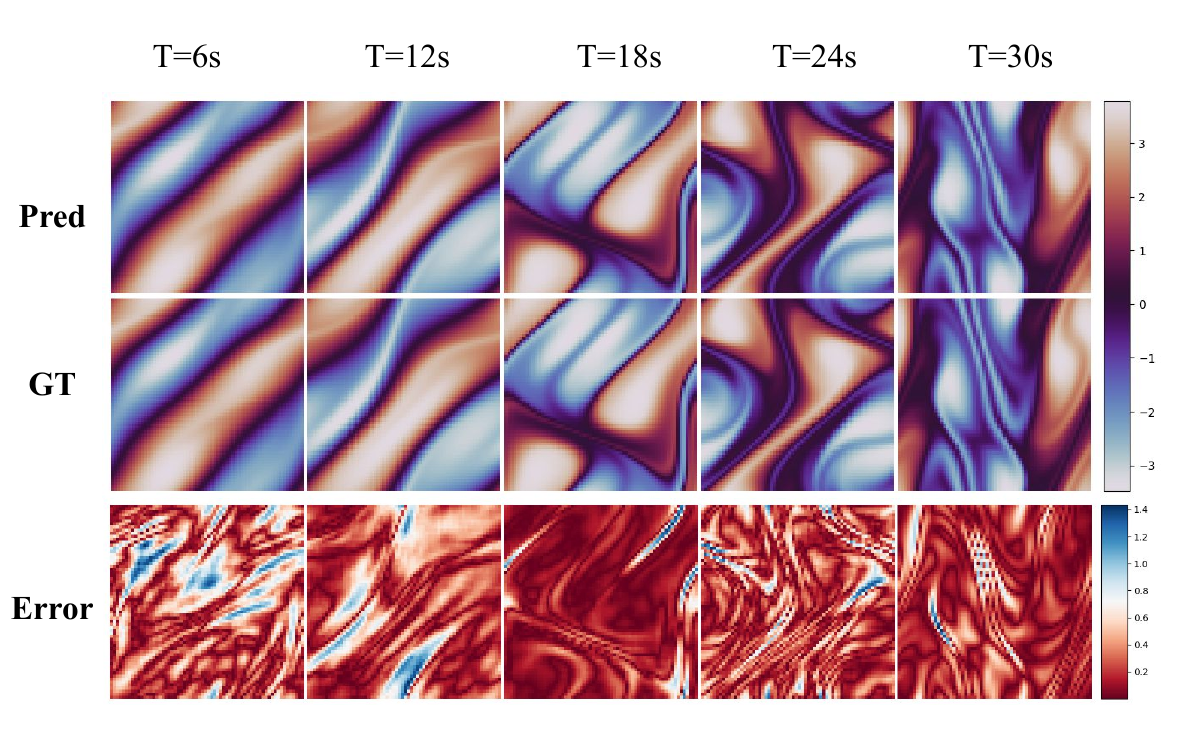}
    \vspace{-3mm}
    \caption{Sample prediction 1 of the vorticity field.}
\end{subfigure}
\begin{subfigure}{\textwidth}
\centering
    \includegraphics[width=0.9\linewidth]{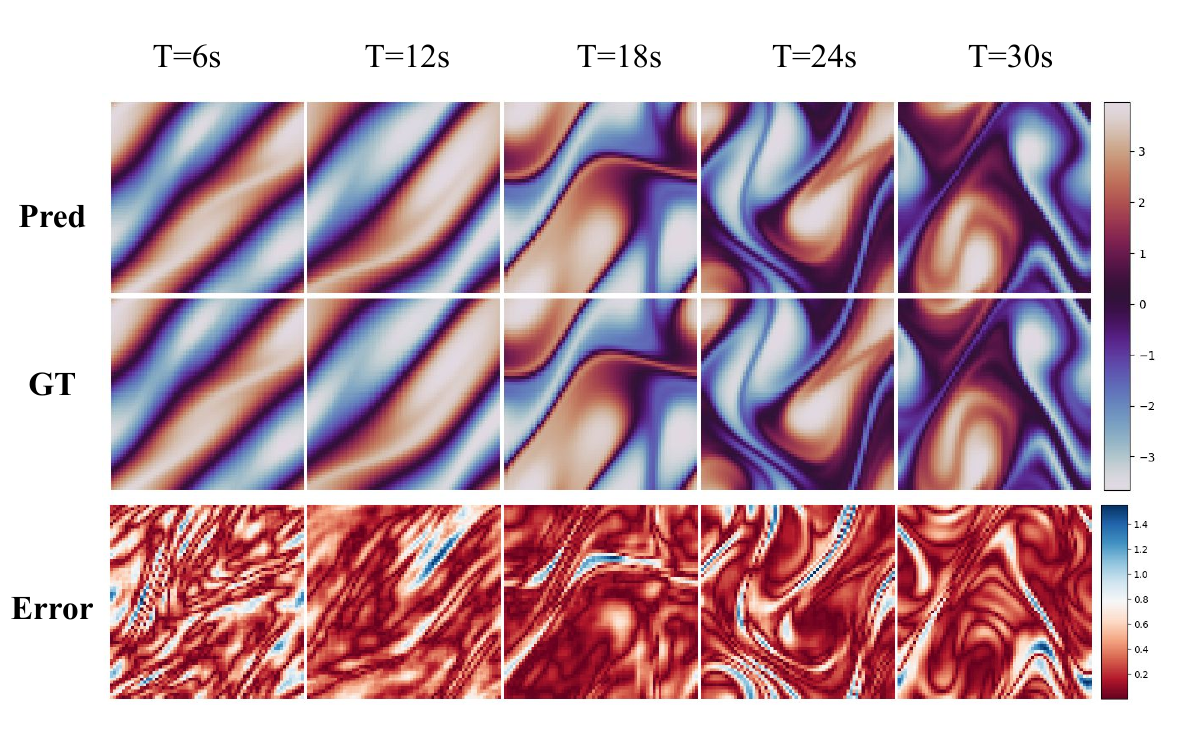}
        \vspace{-3mm}
    \caption{Sample prediction 2 of the vorticity field.}
\end{subfigure}
\end{figure}

\subsection{Tank sloshing}
\begin{figure}[H]
\begin{subfigure}{\textwidth}
    \includegraphics[width=\linewidth]{twophase_varying_height_a.pdf}
    \caption{Sample prediction 1 of the $x$-component of the velocity field $\mathbf{u}$ with liquid surface height: $h=15$. Unit: $m/s$.}
\end{subfigure}
\begin{subfigure}{\textwidth}
    \includegraphics[width=\linewidth]{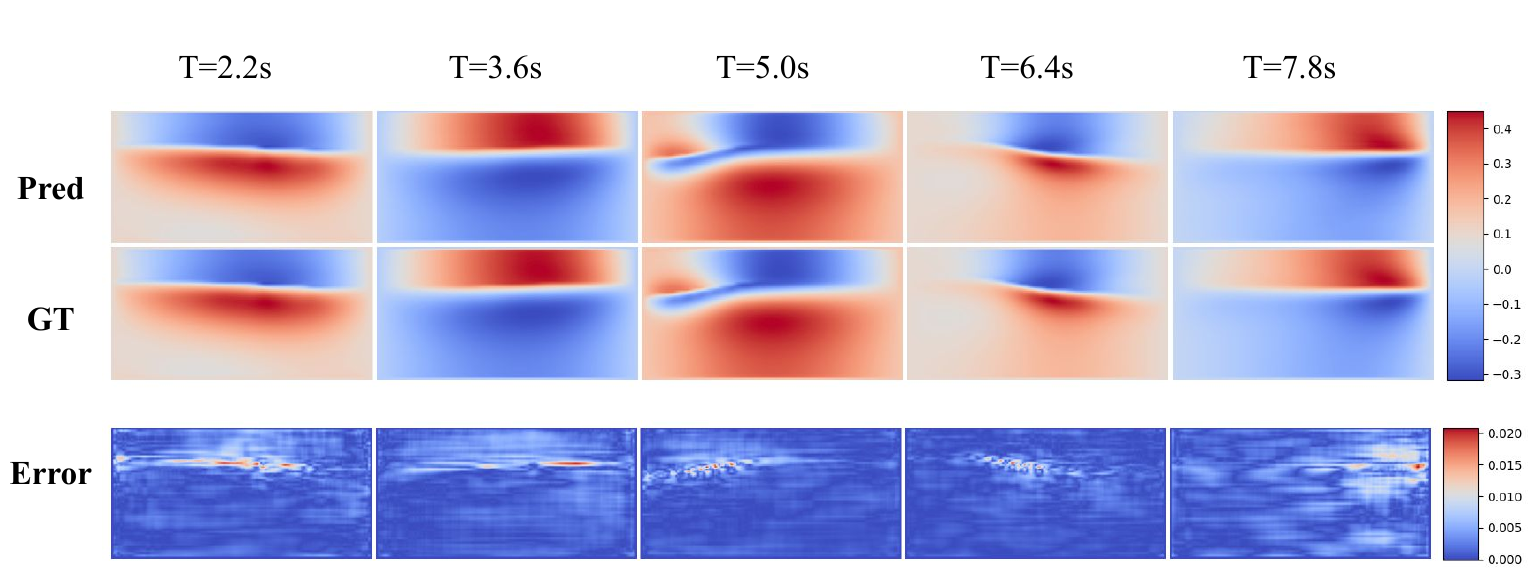}
    \caption{Sample prediction 2 of the $x$-component of the velocity field $\mathbf{u}$ with liquid surface height: $h=44$. Unit: $m/s$.}
\end{subfigure}
\end{figure}
\begin{figure}[h]
\begin{subfigure}{\textwidth}
    \includegraphics[width=\linewidth]{twophase_varying_freq_a.pdf}
    \caption{Sample prediction 1 of the $x$-component of the velocity field $\mathbf{u}$ with oscillation frequency: $\omega=7.12 ~\text{rad/s}$. Unit: $m/s$.}
\end{subfigure}
\begin{subfigure}{\textwidth}
    \includegraphics[width=\linewidth]{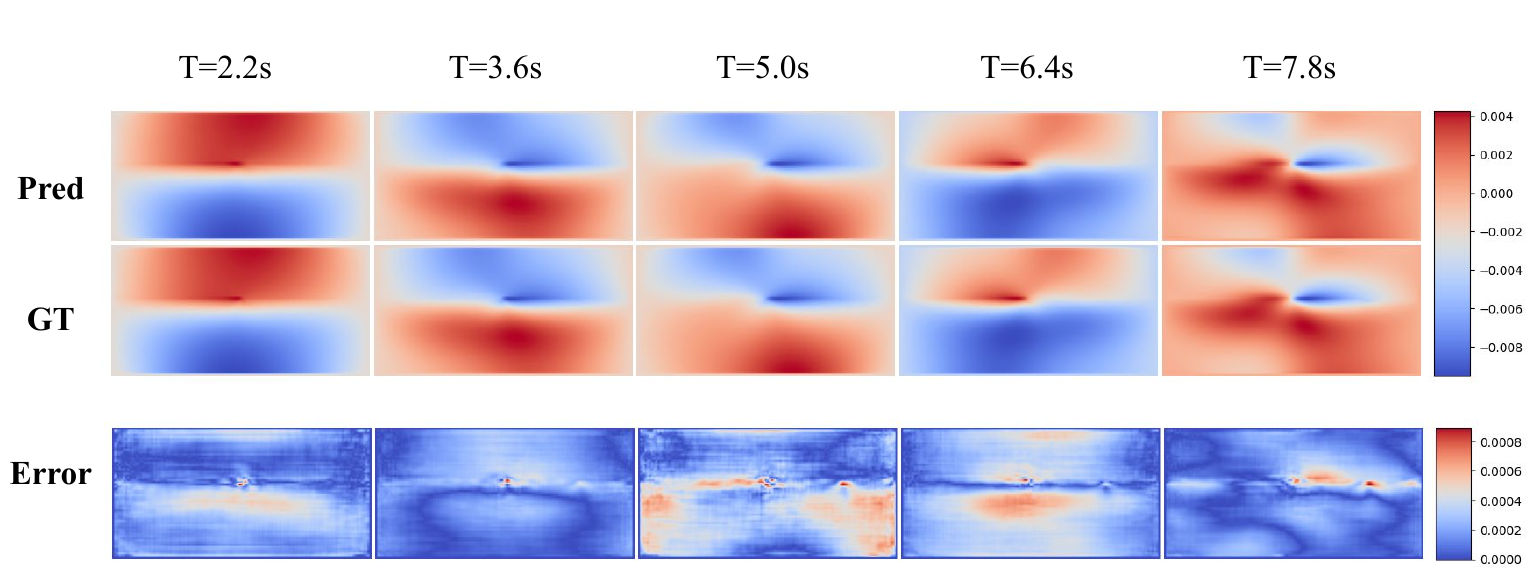}
    \caption{Sample prediction 2 of the $x$-component of the velocity field $\mathbf{u}$ with oscillation frequency: $\omega=1.47 ~\text{rad/s}$. Unit: $m/s$.}
\end{subfigure}
\end{figure}

\subsection{Shallow water}
\begin{figure}[H]
\begin{subfigure}{\textwidth}
    \includegraphics[width=\linewidth]{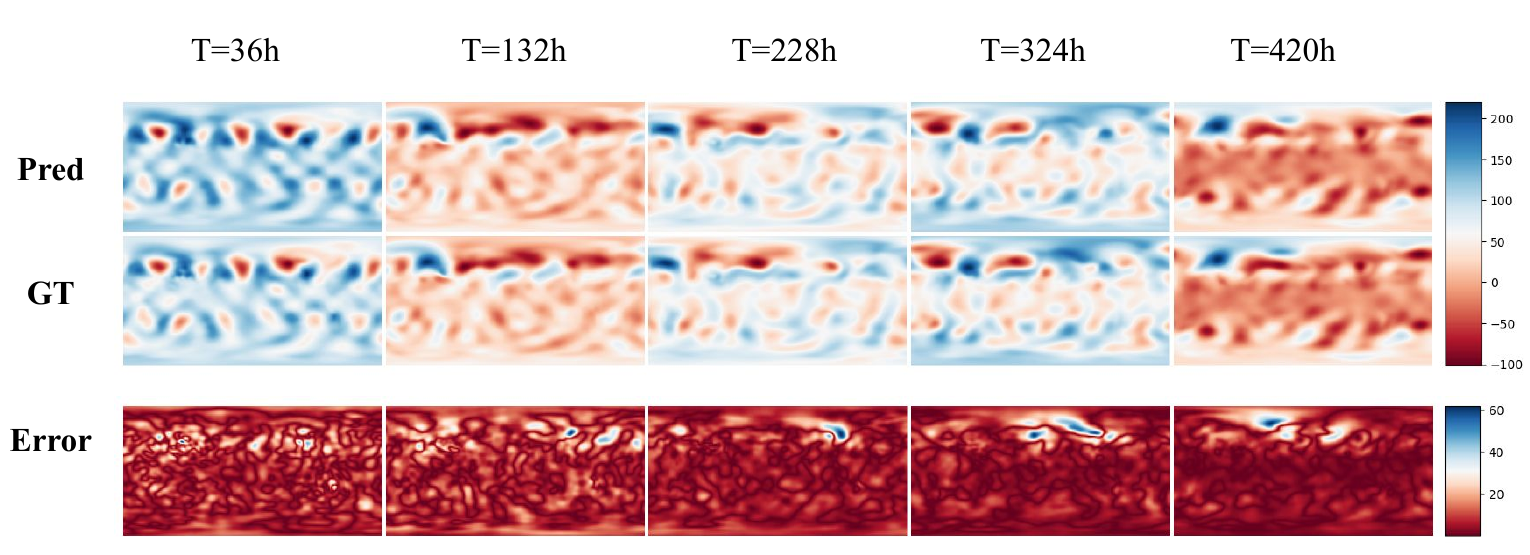}
    \caption{Sample prediction of pressure field $\eta$. Unit: $Pa$.}
\end{subfigure}
\begin{subfigure}{\textwidth}
    \includegraphics[width=\linewidth]{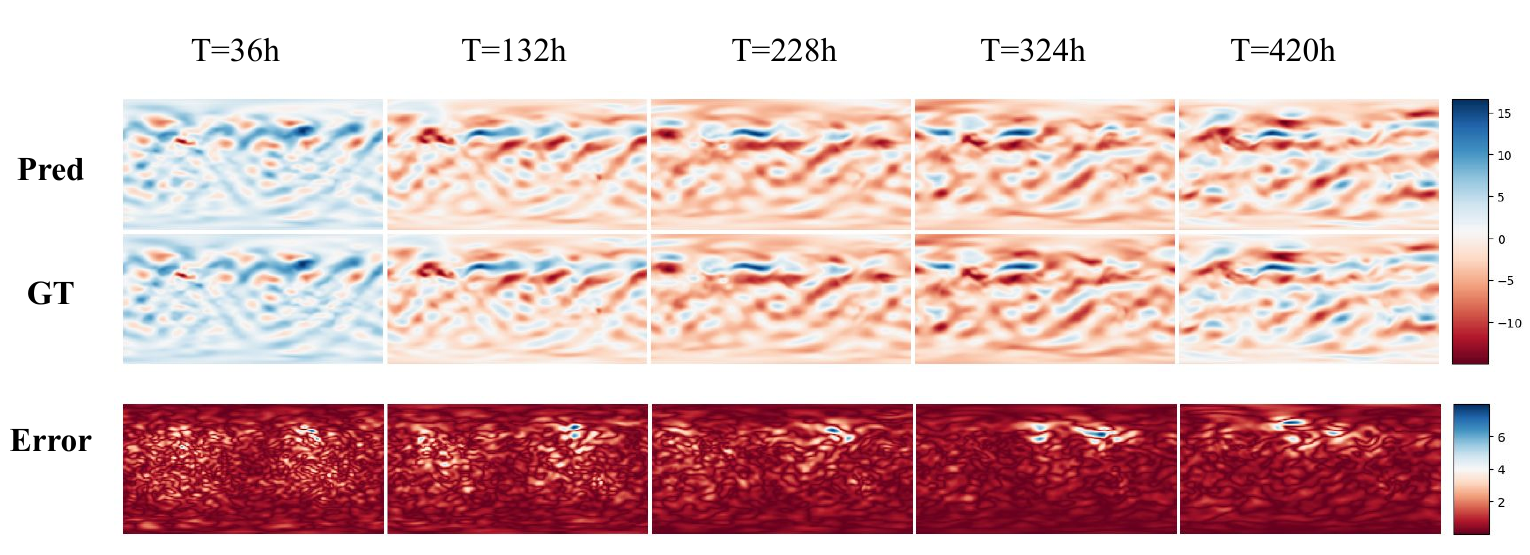}
    \caption{Sample prediction of the $x$-component of velocity field $\mathbf{u}$.. Unit: $m/s$.}
\end{subfigure}
\begin{subfigure}{\textwidth}
    \includegraphics[width=\linewidth]{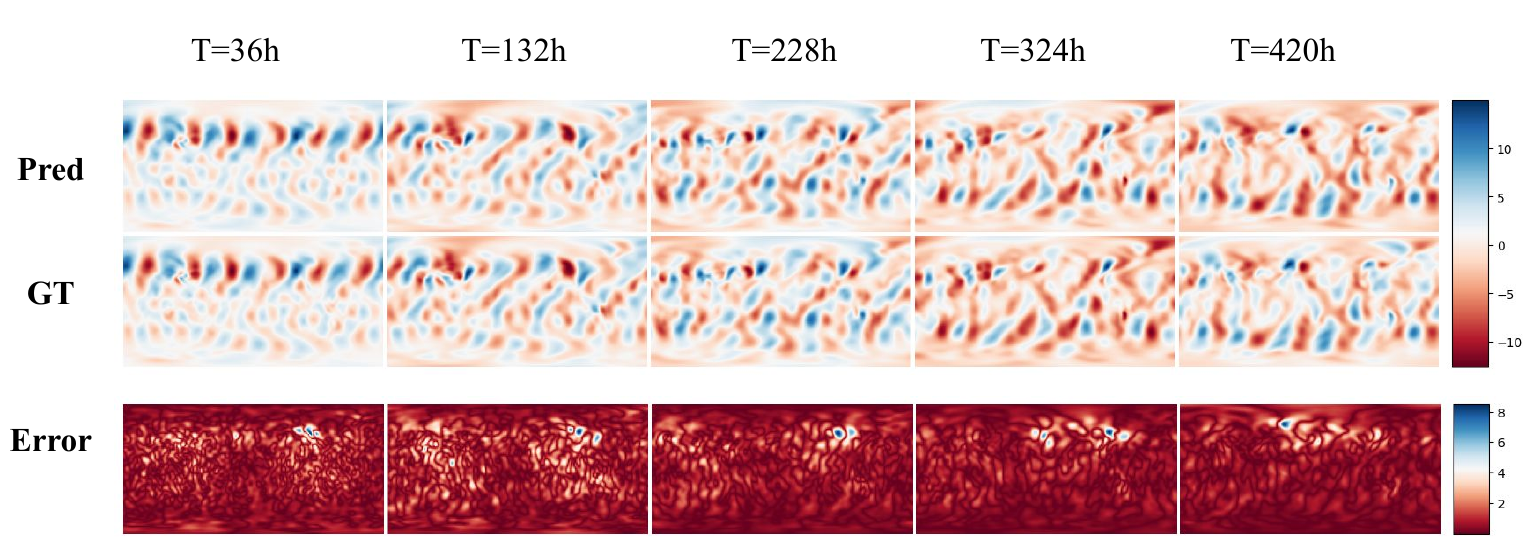}
    \caption{Sample prediction of the $y$-component of velocity field $\mathbf{u}$.. Unit: $m/s$.}
\end{subfigure}
\end{figure}
\vspace{-2mm}
\end{suppinfo}

\end{document}